\documentclass[twocolumn,10pt]{asme2e}

\usepackage{graphicx}

\usepackage{amsmath}
\usepackage{bm}
\usepackage{widetext}
\usepackage{flushend}
\usepackage{cuted}
\usepackage{mathptmx}
\usepackage{array,contour}
\usepackage{float}
\newcommand{\gbf}[1] {\contour[5]{black}{${#1}$}}
\newcommand{\be}{\begin{equation}}
\newcommand{\ee}{\end{equation}}
\newcommand{\beq}{\begin{equation}}
\newcommand{\eeq}{\end{equation}}
\newcommand{\bed}{\begin{displaymath}}
\newcommand{\eed}{\end{displaymath}}
\newcommand{\beqa}{\begin{eqnarray}}
\newcommand{\eeqa}{\end{eqnarray}}
\newcommand{\beqann}{\begin{eqnarray*}}
\newcommand{\eeqann}{\end{eqnarray*}}
\newcommand{\bseq}{\begin{subequations}}
\newcommand{\eseq}{\end{subequations}}

\newcommand{\mat}[2]{\left[ \begin{array}{#1} #2 \end{array} \right] }

\newcommand{\ba}{\begin{array}}
\newcommand{\ea}{\end{array}}

\confshortname{IDETC/CIE 2015}
\conffullname{the ASME 2015 International Design Engineering Technical Conferences \&\\
              Computers and Information in Engineering Conference}

\confdate{August 2-5}
\confyear{2015}
\confcity{Boston}
\confcountry{USA}

\papernum{DETC2015/MR-46292}

\title{Kinematic Analysis and Trajectory Planning of the Orthoglide 5-Axis}
\author{S. Caro, D. Chablat, P. Lemoine, P. Wenger
    \affiliation{
		Institut de Recherche en Communications et 
	Cybern\'etique de Nantes~(IRCCyN) \\
	(UMR CNRS 6597) France\\
    Email addresses: \\ 
		\{Stephane.Caro, Damien.Chablat, Philippe.Lemoine, Philippe.Wenger\}@irccyn.ec-nantes.fr \\
    }	
}

\begin{document}

\maketitle    
\begin{abstract}
{\it The subject of this paper is about the kinematic analysis and the trajectory planning of the Orthoglide 5-axis. The Orthoglide 5-axis a five degrees of freedom parallel kinematic machine developed at IRCCyN and is made up of a hybrid architecture, namely, a three degrees of freedom translational parallel manipulator mounted in series with a two degrees of freedom parallel spherical wrist. The simpler the kinematic modeling of the Orthoglide 5-axis, the higher the maximum frequency of its control loop. Indeed, the control loop of a parallel kinematic machine should be computed with a high frequency, i.e., higher than 1.5 MHz, in order the manipulator to be able to reach high speed motions with a good accuracy. Accordingly, the direct and inverse kinematic models of the Orthoglide 5-axis, its inverse kinematic Jacobian matrix and the first derivative of the latter with respect to time are expressed in this paper. It appears that the kinematic model of the manipulator under study can be written in a quadratic form due to the hybrid architecture of the Orthoglide 5-axis. As illustrative examples, the profiles of the actuated joint angles (lengths), velocities and accelerations that are used in the control loop of the robot are traced for two test trajectories.}
\end{abstract}
\section*{INTRODUCTION}
Parallel kinematics machines become more and more popular in industrial applications.This growing attention is inspired by their essential advantages over serial manipulators that have already reached the dynamic performance limits. In contrast, parallel manipulators are claimed to offer better accuracy, lower mass/inertia properties, and higher structural stiffness (i.e. stiffness-to-mass ratio)~\cite{Merlet:2006}. These features are induced by their specific kinematic architecture, which resists to the error accumulation in kinematic chains and allows convenient actuators location close the manipulator base. Besides, the links work in parallel against the external force/torque, eliminating the cantilever-type loading and increasing the manipulator stiffness. 

Unlike the Variax proposed by Gidding \& Lewis in Chicago in 1994, the delta robot invented by Clavel \cite{Clavel:1988} has known a great success for pick and place applications. One reason for this success is the simplicity of the kinematic and dynamic models compared to the models of the Gough-Stewart platform~\cite{Merlet:2006}. Indeed, the performance of parallel robots may vary considerably within their workspace, which is often small compared to the volume occupied by the machine. It is noteworthy that the inverse kinematics of a parallel manipulator is usually easy to calculate when the actuated joints are prismatic joints as the corresponding equations to be solved are quadratic. However, the inverse Jacobian matrix of such manipulators is more difficult to express and its computing time higher. Several five degrees of freedom~(dof) parallel manipulators have been synthesized in the literature the last few decades. However, their complexity make them difficult to build and use in general. Moreover, the use of fully parallel manipulators leads to robots with five limbs whose mutual collisions or geometric constraints reduce the workspace size.  The {\it Tripteron} is one of the simplest translational parallel robot with three degrees of freedom that can be found in the literature~\cite{Gosselin:2004}. However, this architecture is not suitable for machining operations since its legs are subjected to buckling.

As a consequence, a five dof hybrid machine, named Orthoglide 5-axis, has been developed at IRCCyN. This machine is composed of three dof translational parallel manipulator, named Orthoglide 3-axis, mounted in series with a two dof spherical parallel manipulator, named Agile Eye 2-axis. The Orthoglide 3-axis has the advantages of both serial and parallel kinematic architectures such as regular workspace, homogeneous performances, good dynamic performances and stiffness. The interesting features of the Orthoglide 3-axis are large regular dextrous workspace, uniform kinetostatic performances, good compactness \cite{Pashkevich:2005} and high stiffness \cite{Pashkevich:2008}. Besides, the translational and rotational motions of the end-effector (tool) are partially decoupled with the hybrid architecture of the Orthoglide 5-axis.

This paper is organized as follows. The next section deals with the kinematic modeling of the Orthoglide 5-axis. Then, some trajectories are generated using a simplified computed torque control loop and tested experimentally. Finally, some conclusions and future work are presented.
\section*{THE ORTHOGLIDE 5-AXIS}
\subsection*{A hybrid architecture}
\begin{figure}[!ht]
  \begin{center}
        \includegraphics[scale=0.8]{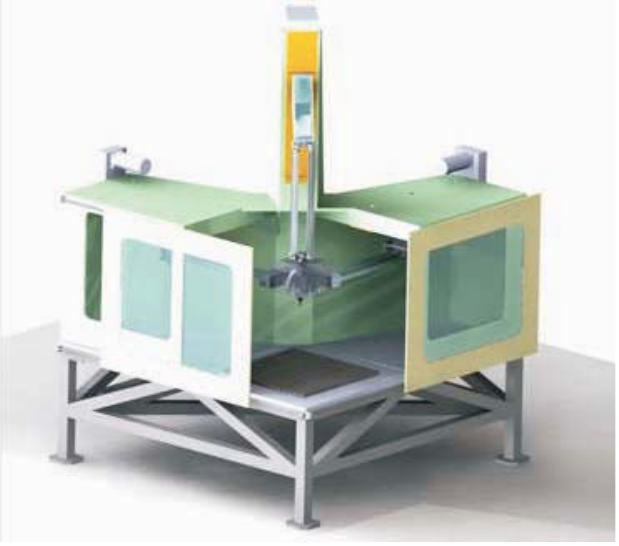}
        \caption{Digital mock-up of the Orthoglide 5-Axis}
        \protect\label{figure:robot_complet}
  \end{center}
\end{figure}

\begin{figure}[!ht]
  \begin{center}
        \includegraphics[scale=0.25]{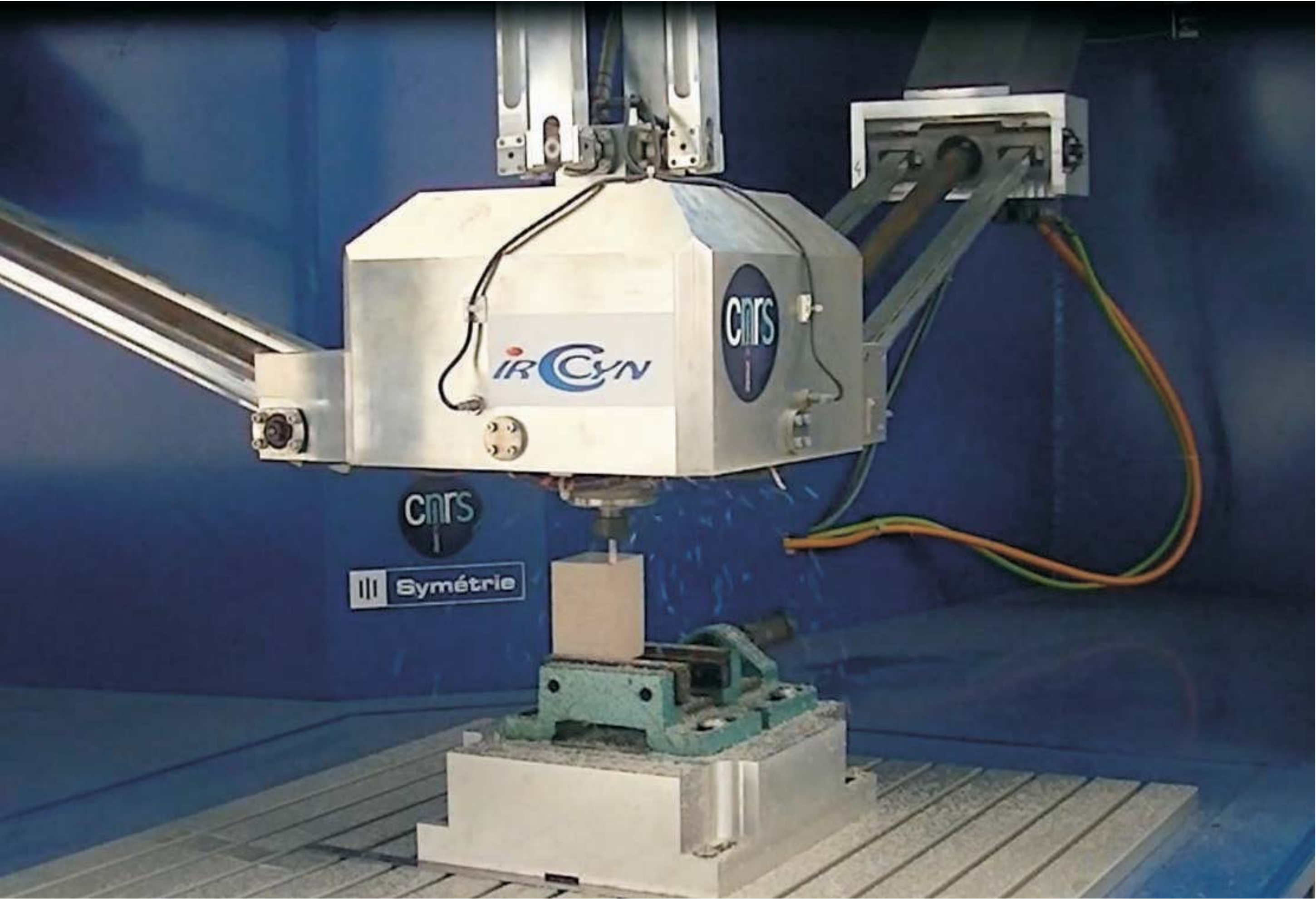}
        \caption{Semi industrial prototype of the Orthoglide 5-Axis performing a machining operation}
        \protect\label{figure:robot}
  \end{center}
\end{figure}

Figure \ref{figure:robot_complet} depicts a CAD modeling of the Orthoglide 5-axis and figure \ref{figure:robot} shows a semi industrial prototype of the Orthoglide 5-axis located at IRCCyN. The Orthoglide 5-axis is a hybrid parallel kinematics machine (PKM) composed of a 3-dof translational parallel manipulator, the Orthoglide 3-axis, mounted in series with two dof parallel spherical manipulator, the Agile Eye  2-axis.  The Agile Eye  2-axis is spherical wrist developed at Laval University~\cite{Gosselin:1994}. The architecture of the Orthoglide 5-axis was presented in \cite{Ur-Rehman:2008} as well as in~\cite{Chablat:2006}.
\subsection*{Orthoglide 3-Axis}
The Orthoglide 3-axis is composed of three identical legs. Each leg is made up of a prismatic joint, a revolute joint, a parallelogram joint and another revolute joint. The first joint, i.e., the prismatic joint of each leg, is actuated and the end-effector is attached to the other end of each leg. Hence, the Orthoglide 3-axis is a PKM with movable foot points and constant chain lengths.

The kinematics of the Orthoglide 3-axis was defined in \cite{Wenger:2000}. Let ${\gbf{ \rho}}=[\rho_1~\rho_2~\rho_3]^t$ denote  the vector of linear joint variables and ${\bf p}= [x~y~z]^t$ denote the Cartesian coordinate vector of the position of the end-effector. The loop closure of the Orthoglide 3-axis leads to the following three constraint equations: 
\begin{eqnarray}
\left( x - \rho_1 \right) ^{2}+{y}^{2}+{z}^{2}&=& l_1^{2}  \\
{x}^{2}+ \left( y- \rho_2 \right) ^{2}+{z}^{2}&=& l_2^{2}  \\
{x}^{2}+{y}^{2}+ \left( z- \rho_3 \right) ^{2}&=& l_3^{2}
\label{eq:system}
\end{eqnarray}
Therefore, the inverse Jacobian matrix ${\bf J}^{-1}_{O}$ of the Orthoglide 3-axis takes the form:
\begin{equation}
{\bf J}^{-1}_{O}= 
\left[ \begin {array}{ccc} 
1&-{\dfrac{y}{\rho_1-x}} &-{\dfrac{z}{\rho_1-x}}\\ 
-{\dfrac{x}{\rho_2-y}}&1&-{\dfrac{z}{\rho_2-y}}\\ 
-{\dfrac{x}{\rho_3-z}}&-{\dfrac{y}{\rho_3-z}}&1\end {array} \right] 
\end{equation}
It appears that the Orthoglide 3-axis can have up to two assembly modes, i.e., two solutions to its direct geometric model, and up to eight working modes, i.e., eight solutions to its inverse geometric model. Moreover, those solutions can be easily obtained by solving simple quadratic equations~\cite{Pashkevich:2006}. The lengths of the parallelogram joints and the joint limits were obtained by using the method presented in~\cite{Chablat:2003} in order the manipulator to get a cube-shaped translational workspace of 500~mm edge. 
\subsection*{Agile Eye 2-axis}
\begin{figure}[!ht]
  \begin{center}
        \includegraphics[scale=0.5]{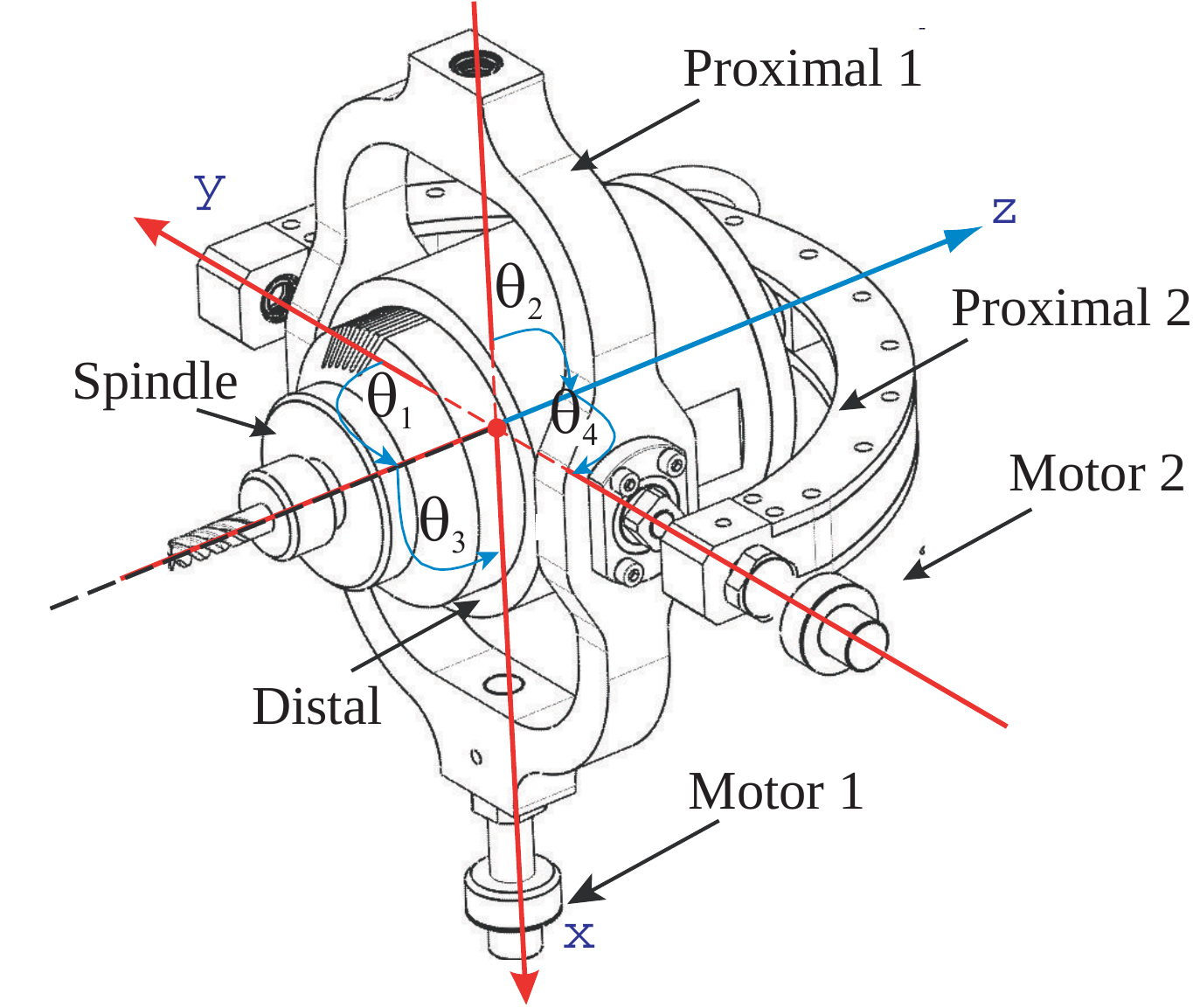}
        \caption{Spindle mounted on the two dof spherical wrist}
        \protect\label{figure:Wrist}
  \end{center}
\end{figure}

Figure~\ref{figure:Wrist} illustrates a spindle mounted on the two dof spherical parallel manipulator, the latter being mounted in series on the Orthoglide 3-axis in order to get the Orthoglide 5-axis.
The Agile Eye 2-axis, namely the two dof spherical parallel manipulator, consists of a closed kinematic chain composed of five components: the proximal~1 link, the proximal~2 link, the distal link, the terminal link and the base. These five links are connected by means of revolute joints, the two revolute joints connected to the base being actuated. Let us notice that all revolute joints axes intersect. The shape of the proximal and distal links were determined in~\cite{Chablat:2006b} in order the wrist to have a high stiffness.

The kinematics of the Agile eye was defined in \cite{Gosselin:1994}. Note that the orientation workspace of the Orthoglide 5-axis is limited to $\pm 45$ degrees due to mechanical constraints. For machining operations, the orientation of the tool is defined by two rotation angles~$\alpha$ and~$\beta$ rotating respectively around the $x$ and $y$~axes. A IBAG spindle HFK 95.1 is mounted on the spherical wrist. Its power is equal to $1$~KW and its maximum speed is equal to 42000~rpm. The distance~$l$ between the tool tip and the geometric and rotation center of the wrist should be higher than $72$~mm~($l>72$mm).

The geometric, kinematic and dynamic models of the Agile Eye 2-axis are given in~\cite{Caron:1997}. Here, its inverse Jacobian matrix is recalled in order to compute the velocity of the geometric center of the wrist with respect to the tool tip and the corresponding actuated prismatic joint rates of the Orthoglide 3-axis. 

The Agile Eye admits two assembly modes that are easily obtained and discriminated by solving a quadratic equation. The Cartesian coordinate vector~${\bf p}_{tt}$ of the tool tip is expressed as a function of angles~$\alpha$ and~$\beta$ as follows:
\begin{equation}
{\bf p}_{tt}=\left[ \begin {array}{c} 
-\sin \left( \beta \right) \\ 
\sin \left( \alpha \right) \cos \left( \beta \right) \\ 
-\cos \left( \alpha \right) \cos \left( \beta \right) 
\end {array} \right] 
\end{equation}
The actuated joint angles~$\theta_1$ and $\theta_2$ are expressed as a function of angles~$\alpha$ and~$\beta$ as follows:
\begin{eqnarray}
\theta_1&=& -\arctan \left(\dfrac{-\sin(\alpha) \cos(\beta)}{\cos(\alpha)\cos(\beta)}\right) \\
\theta_2&=& \arctan \left(\dfrac{\sin(\beta)}{\cos(\alpha)\cos(\beta)}\right) 
\end{eqnarray}

The inverse kinematic Jacobian matrix, ${\bf J}^{-1}_{A}$, of the parallel spherical manipulator is expressed as:
\begin{equation}
{\bf J}^{-1}_{A}=
 \left[ \begin {array}{cc} 1& 0 \\
 \dfrac{S\theta_2 S\theta_1 C\beta}{S\beta S\theta_2 + C\theta_1 C\beta C\theta_2} &
 \dfrac{S\beta C\theta_1 S\theta_2 + C\theta_2 C\beta}{S\beta S\theta_2 + C\theta_1 C\beta C\theta_2}
\end {array} \right]
\end{equation}
where $C\theta_i=\cos(\theta_i$), $S\theta_i=\sin(\theta_i$), $C\alpha=\cos(\alpha)$, $S\alpha=\sin(\alpha)$, $C\beta=\cos(\beta)$, $S\beta=\sin(\beta)$ and

\begin{equation}
	 \mat{c}{\dot\theta_1 \\ \dot\theta_2} = {\bf J}^{-1}_{A} \, \mat{c}{\dot\alpha \\ \dot\beta}
\end{equation}
The relation between the Cartesian velocities of the geometric center of the spherical wrist and the output angle rates is given as a function of the tool length~$l$ by:
\begin{equation}
	 \mat{c}{\dot x \\ \dot y \\ \dot z} = {\bf J}^{-1}_{C} \, \mat{c}{\dot\alpha \\ \dot\beta}
\end{equation}
with
\begin{equation}
{\bf J}^{-1}_{C}=
\left[ \begin {array}{cc} {\dfrac{y l C\alpha  C\beta}{-x+{ \rho_1}}}+{\dfrac{z l S\alpha C\beta }{-x+{ \rho_1}}}&
 l C\beta -{\dfrac{y l S\alpha S\beta }{-x+{ \rho_1}}}+{\dfrac{z l C\alpha S\beta }{-x+{ \rho_1}}}\\ 
- l C\alpha C\beta +{\dfrac{z l S \alpha C\beta}{-y+{ \rho_2}}}&
-{\dfrac{x l C\beta }{-y+{ \rho_2}}}+ l S\alpha S\beta +{\dfrac{z l C\alpha S\beta }{-y+{ \rho_2}}}\\
 {\dfrac{y l C\alpha  C\beta}{-z+{ \rho_3}}}- l S\alpha C\beta &
-{\dfrac{x l C\beta }{-z+{ \rho_3}}}-{\dfrac{y l S\alpha S\beta}{-z+{ \rho_3}}}- l C\alpha S\beta 
\end {array} \right] 
\end{equation}
\subsection*{Coupling two parallel robots}
The assembly of two parallel robots in series leads to a simple kinematic model of the Orthoglide 5-axis. Indeed, the direct kinematics of the latter is obtained by solving first the direct kinematics of the Orthoglide 3-axis in order to get the Cartesian coordinates of the geometric center of the spherical wrist and then by calculating the direct kinematics of the spherical wrist in order to express the pose of the tool. About the inverse kinematics of the Orthoglide 5-axis, the inverse of the spherical wrist is first solved, then the inverse kinematics of the Orthoglide 3-axis is solved. It is noteworthy that a translational motion of the geometric center of the wrist does not change the orientation of the tool. However, a rotation of the tool about its end leads to translational displacement of the geometric center of the spherical wrist and as a result to actuated prismatic joint displacements.
The inverse kinematic modeling of the Orthoglide 5-axis is expressed as:
\begin{equation}
\dot{\bf q} = {\bf J}^{-1} {\bf t}
\end{equation}
where $\dot{\bf q}= \left[\dot{\theta_1}~\dot{\theta_2}~\dot{\rho_1}~\dot{\rho_1}~\dot{\rho_1}\right]^T$ and ${\bf t}= \left[ \dot{\alpha}~\dot{\beta}~\dot{x}~\dot{y}~\dot{z}\right]^T$.
We can describe the inverse Jacobian matrix as three sub-matrices coming from the both parallel robots. 
\begin{equation}
{\bf J}^{-1}= \left[
\begin{array}{cccc}
{\bf J}^{-1}_A & \bf{0}_{2 \times 3} \\
{\bf J}^{-1}_O {\bf J}^{-1}_C & {\bf J}^{-1}_O
\end{array}
\right]
\end{equation}
as is written in \ref{equation:jacobien_complet}.
\begin{figure*}
\begin{equation}
 {\bf J}^{-1}= \left[ \begin {array}{ccccc} 1&0&0&0&0\\ 
{\dfrac{S\theta_2 S\alpha C\beta}{S\beta S\theta_2 
+ C\alpha C\beta C\theta_2 }}&
{\dfrac{ C\alpha S\beta S\theta_2  + C\theta_2 C\beta  }{S\beta S\theta_2  + C\alpha C\beta C\theta_2 }}
&0&0&0\\ 
{\dfrac{y l C\alpha C\beta }{-x+{\rho_1}}}+{\dfrac{z l S \alpha C\beta }{-x+{\rho_1}}}& 
l C\beta -{\dfrac{y l S\alpha S\beta}{-x+\rho_1}}+{\dfrac{z l C\alpha S \beta }{-x+\rho_1}}&
1&
-{\dfrac{y}{-x+{\rho_1}}}&-{\dfrac{z}{-x+{\rho_1}}}\\ 
-l C\alpha C \beta +{\dfrac{z l S\alpha C\beta }{-y+{\rho_2}}}&-{\dfrac{x l C\beta }{-y+{\rho_2}}}+ l S \alpha S\beta +{\dfrac{z l C\alpha S\beta }{-y+{\rho_2}}}&
-{\dfrac{x}{-y+{\rho_2}}}&1&
-{\dfrac{z}{-y+{\rho_2}}}\\ 
{\dfrac{y l C\alpha C\beta}{-z+{\rho_3}}}- l S\alpha C\beta &
-{\dfrac{x l C\beta}{-z+{\rho_3}}}-{
\dfrac{y l S\alpha S\beta}{-z+{\rho_3}}}- l C\alpha S\beta  &
-{\dfrac{x}{-z+{\rho_3}}}&
-{\dfrac{y}{-z+{\rho_3}}}&1
\end {array}
 \right]
\label{equation:jacobien_complet}
\end{equation}
\end{figure*}
Note that the first derivative of $\bf {J}^{-1}$ with respect to time can be easily obtained by using a symbolic computation software. However, its expression is too lengthy to be displayed in the paper.
\section*{TRAJECTORY GENERATION}
The trajectory planning approach described in~\cite{Khalil:2002} has been used for the trajectory generation of the Orthoglide 5-axis. Note that the trajectories are defined in the task space, i.e., the robot workspace, whereas the control loop requires data in the joint space. According the design rules presented in \cite{Chablat:2003}, the boundaries of the Orthoglide 3-axis workspace are defined in the Cartesian space. Therefore, the direct kinematics is computed at the frequency of the control loop in order to check whether the moving-platform of the robot is within its Cartesian workspace at anytime or not. 

Here, the system dynamics is decoupled in order to consider five separate actuators moving an equivalent mass and analyze the stability of the system. In the robot control scheme, the dynamic effects have been taken into account in order to improve the classical PID control scheme. Therefore, the torque~$\Gamma_R$ associated to the rotational motions and the torque~$\Gamma_T$ associated to the translational motions are expressed as follows:
\begin{eqnarray}
\Gamma_R&=& J\left( \ddot{\theta_i} + {K}_{P_R} (\theta^d_i - \theta_i)+{K}_{D_R} (\dot{\theta}^d_i - \dot{\theta}_i)+{K}_{I_R} \int_{t0}^t {(\theta^d_i - \theta_i)}\right) \nonumber \\
\Gamma_T&=& M\left( \ddot{\rho_j} + {K}_{P_T} (\rho^d_j - \rho_j)+{K}_{D_T} (\dot{\rho}^p_j - \dot{\rho}_j)+{K}_{I_T} \int_{t0}^t {(\rho^d_j - \rho_j)} \right)\nonumber 
\end{eqnarray}
for $i=1..2$, $j=1..3$, $J=0.2772~Kg.m^2$, $M=91.6278~Kg$, $\theta^d_i$ and $\dot{\theta}^d_i$ denote the prescribed actuated joint angles and rates. $\theta_i$ and $\dot{\theta}_i$ denote the actual joint angles and rates. The values of coefficients~$J$ and $M$ mainly depend on the motor inertia, the gears and the axis of the ball screws. Note that the effect of the inertia and mass of the elements in motion is somehow the corresponding inertia and mass divided by the square of the gearhead ratio.
As a result, for the rotational and translational parts, we obtain:
\begin{eqnarray}
\omega&=&   49~rad/s \nonumber \\
K_{P_R}&=& K_{P_T}= 3 \omega^2=19200		\nonumber \\
K_{V_R}&=& K_{V_T}= 3 \omega=240			\nonumber \\
K_{I_R}&=& K_{I_T}= \omega^3=512000
\end{eqnarray}
where $\omega$ is a function of the torque constant, the dielectric constant, the motor efficiency and its inertia.
\begin{figure}[!ht]
  \begin{center}
        \includegraphics[scale=1.0]{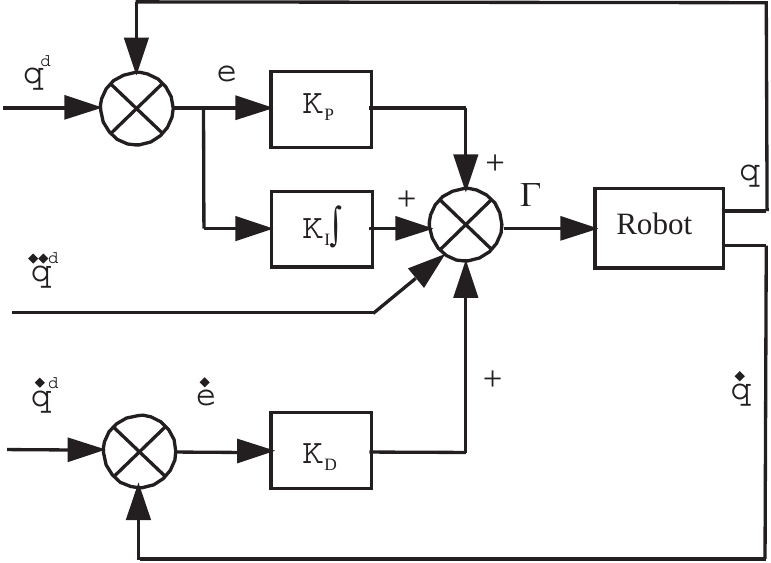}
        \caption{Control scheme of the Orthoglide 5-Axis}
        \protect\label{figure:control_loop}
  \end{center}
\end{figure}
Figure~\ref{figure:control_loop} shows the control scheme of the Orthoglide 5-Axis.
        
\subsection*{Control hardware and control loop}
The control hardware of the Orthoglide 5-axis is a 1103 DSPACE card \cite{Dspace:2015} with a  933~MHz PowerPC. The actuator positions are acquired with a frequency equal to 9~kHz and a 200~Hz low pass filter  is used to compute the actuator velocities. The robot motions are controlled thanks to a sub program working at 1.5~KHz. This  program manages the control loop, the input/output relation and the security system. For instance, if the positional error of the end-effector is larger than 3~mm, the machine will shutdown. Another level of security is on the actuators with electrical sensors stroke end. If enabled, neither the PC nor the Dspace card operates, it amounts to the cable security. The robot trajectories are calculated using Matlab and sent to the Dspace card via an optical fiber. In the Dspace card, the inverse Jacobian matrix is used when the robot is close to its workspace boundaries in order to know whether the Cartesian displacements of the end-effector lead to exceeding joint limits or not.


A pneumatic compensator is mounted along the vertical axis in order to reduce the gravity effect. This compensator continuously generates an effort equivalent to the weight of the mobile platform, but in the opposite direction. The pressure in the cylinders is controlled by a purely mechanical system. 

Two types of motor are used for the actuated joints.Two Harmonic Drive FFA-20-80-H-E2048 are used to actuate the two-dof parallel spherical wrist and three Parvex Brushless NX430EAF coupled with Kinetic TDU 200 ball screws are used to actuate the three-dof translational parallel manipulator. As a consequence, the Orthoglide 5-axis can reach linear velocities up to 1~m/s and a linear accelerations up to 1~G.

Note that the maximum angular velocity of the wrist motors is equal to $3.27$~rad/s and their maximum angular accelerations is equal to $270$~rad/s$^2$. The maximum linear velocity of the prismatic joints is equal to $1.2$~m/s due to mechanical constraints and their maximum acceleration is equal to $13$ m/s$^2$ due to the value of the maximum continuous torques.

\subsection*{Trajectories}
Three types of trajectories can be realized by the Orthoglide 5-axis: 
\begin{enumerate}
	\item The first trajectory is a linear interpolation between two postures defined by two angles $\alpha$ and $\beta$ and by three Cartesian coordinates $x$, $y$ and $z$ for which we can specify a velocity smaller than the maximum velocity, namely, a ratio of the maximum velocity. 
	\item The second trajectory is a circle defined by a center point, a radius and a percentage of the maximum speed. The orientation of the spindle is changing linearly around the circle.
	\item The last trajectory is described by a G-code file. So far, only G00 and G01 functions have been implemented. The trajectory planner aims to keep the velocity between two postures constant. Some corners were added in order to smooth the path and avoid robot breaks. The radii of corner curvature are defined in order to use acceleration capacity of the machine optimally and to minimize robot speed reduction.
\end{enumerate}

For all trajectories, a linear trajectory is automatically added between the current pose of the robot and its the starting point of the trajectory. Another linear trajectory between the final point of the trajectory and the current pose in order to close the loop. For safety reason, the actual speed is limited to 10\% of the maximal speed during all trajectories.

For the first two types of trajectory, a fifth degree polynomial equation is used to define the position $r$, the velocity $\dot{r}$ and the acceleration $\ddot{r}$ along a linear interpolation:
\begin{eqnarray}
r(t)&=& 10 \left(\dfrac{t}{t_f}\right)^3 - 15 \left(\dfrac{t}{t_f} \right)^4+6\left( \dfrac{t}{t_f}\right)^5 \\
\dot{r}(t)&=&30 \left({\dfrac{t^2}{t_f^{3}}}\right)-60\left(\dfrac{t^{3}}{t_f^{4}}\right)+30\left(\dfrac{t^4}{t_f^5}\right) \\
\ddot{r}(t)&=&60 \left({\dfrac{t}{{t_f}^{3}}}\right)-180 \left({\dfrac{{t}^{2}}{{t_f}^{4}}}\right)+120 \left({\dfrac{{t}^{3}}{{t_f}^{5}}}\right)
\end{eqnarray}

A computed torque control in the joint space is used. Therefore, the trajectory to be followed is projected into the robot joint space by using the inverse geometric model of the Orthoglide 5-axis, the inverse kinematic Jacobian matrix~${\bf J}^{-1}$ and its first time derivative~$\dot{\bf J}^{-1}$, in order to know whether the maximum motor velocities and accelerations are reached or not.

\subsection*{Linear trajectory}
The computing traveling time depends on the total distance between the two robot poses. The minimum traveling time $t_f$ is defined as:
\begin{equation}
t_f = \max\left( \dfrac{|D_T| }{k_{V_T}},\dfrac{|D_R|}{k_{V_R}}\right)
\end{equation}
where $|D_T|$ is the norm of the Cartesian motion,  $|D_R|$ is the norm of the angular motion, $k_{V_T}$ and $k_{V_R}$ are the maximum linear and angular velocities. We can notice that these parameters are defined in the Cartesian workspace and will be larger if the maximum joint velocity and acceleration are exceeded. 
Let $\bf{P}_1$ and $\bf{P}_2$ be the vectors defining the first pose and second pose of the robot end-effector, respectively. Any intermediate pose~$\bf P$ is defined as:
\begin{equation}
\bf{P}(t)= \bf{P}_1 + (\bf{P}_2-\bf{P}_1) r(t)
\end{equation}
The velocity vector $\bf{V}(t)$ is expressed as:
\begin{equation}
\bf{V}(t)= \dot{r}(t) (\bf{P}_2-\bf{P}_1) 
\end{equation}
The acceleration vector $\bf{A}(t)$ takes the form:
\begin{equation}
\bf{A}(t)= \ddot{r}(t) (\bf{P}_2-\bf{P}_1) 
\end{equation}

The corresponding joint coordinate vector is obtained by:
\begin{equation}
\bf{q}(t)= f(\bf{P}(t))
\end{equation}
with ${\bf f}$ denotes the inverse geometric model of the Orthglide 5-axis. The joint velocities are computed with the inverse kinematic Jacobian matrix~${\bf J}^{-1}$ as:
\begin{equation}
\dot{\bf{q}}(t)= {\bf J}^{-1} \bf{V}(t)
\end{equation}
The joint acceleration by using the inverse Jacobian matrix and its first derivation $\dot{\bf J}^{-1}$
\begin{equation}
\ddot{\bf{q}}(t)= {\bf J}^{-1} A(t) + \dot{\bf J}^{-1} V(t)
\end{equation}

Table~\ref{table:ligne} defines a trajectory where the maximum velocity and acceleration are requested between pose~$\bf{P}_2$ and pose~$\bf{P}_3$ (1.2~m/s and 1.2~G).  

\begin{figure}[!ht]
  \begin{center}
	\begin{minipage}[c]{.49\linewidth}
        \includegraphics[scale=0.35]{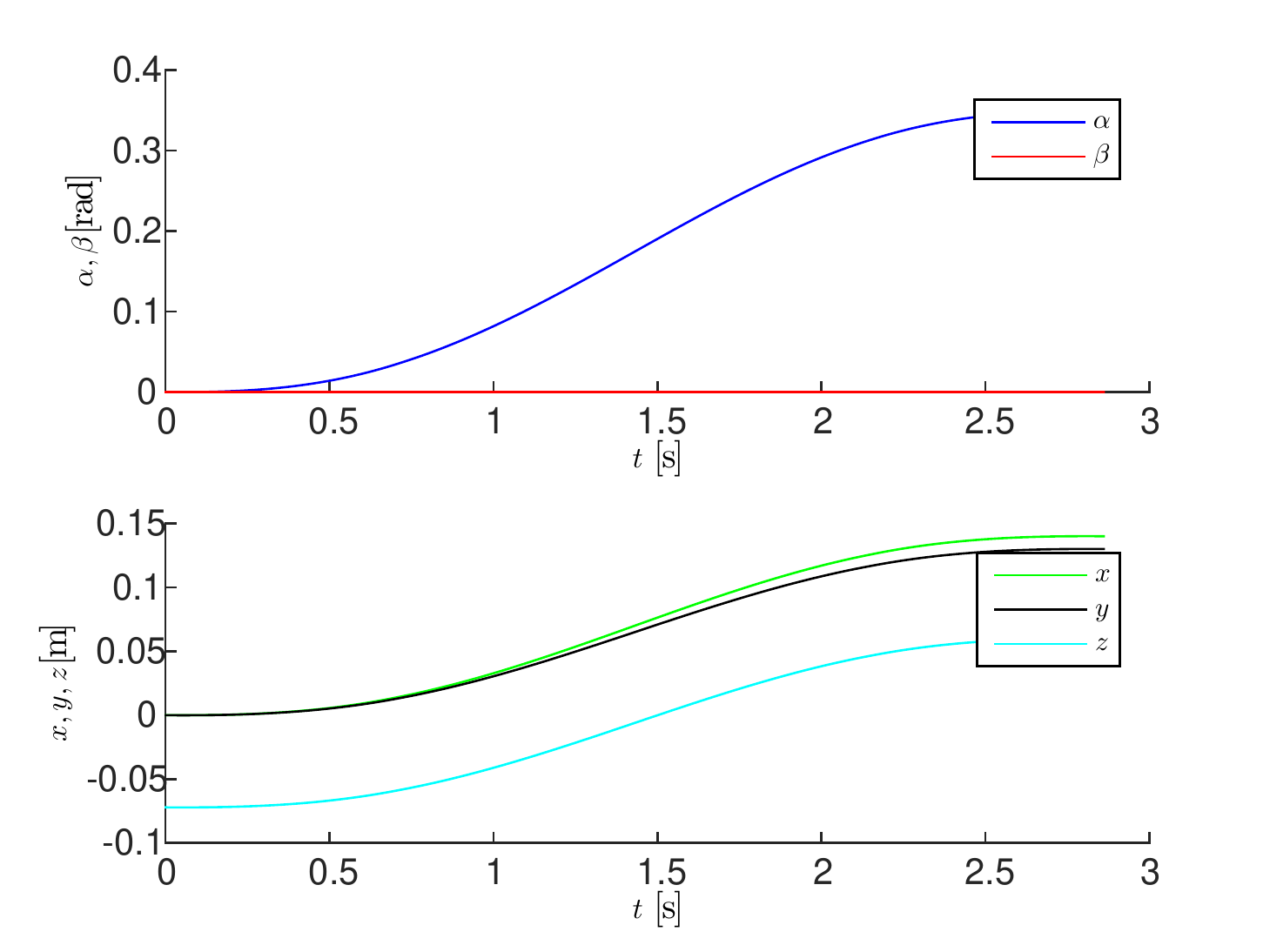} \\
				\small{(a) $\bf{P}_1$ to $\bf{P}_2$}
   \end{minipage} \hfill
   \begin{minipage}[c]{.49\linewidth}        
        \includegraphics[scale=0.35]{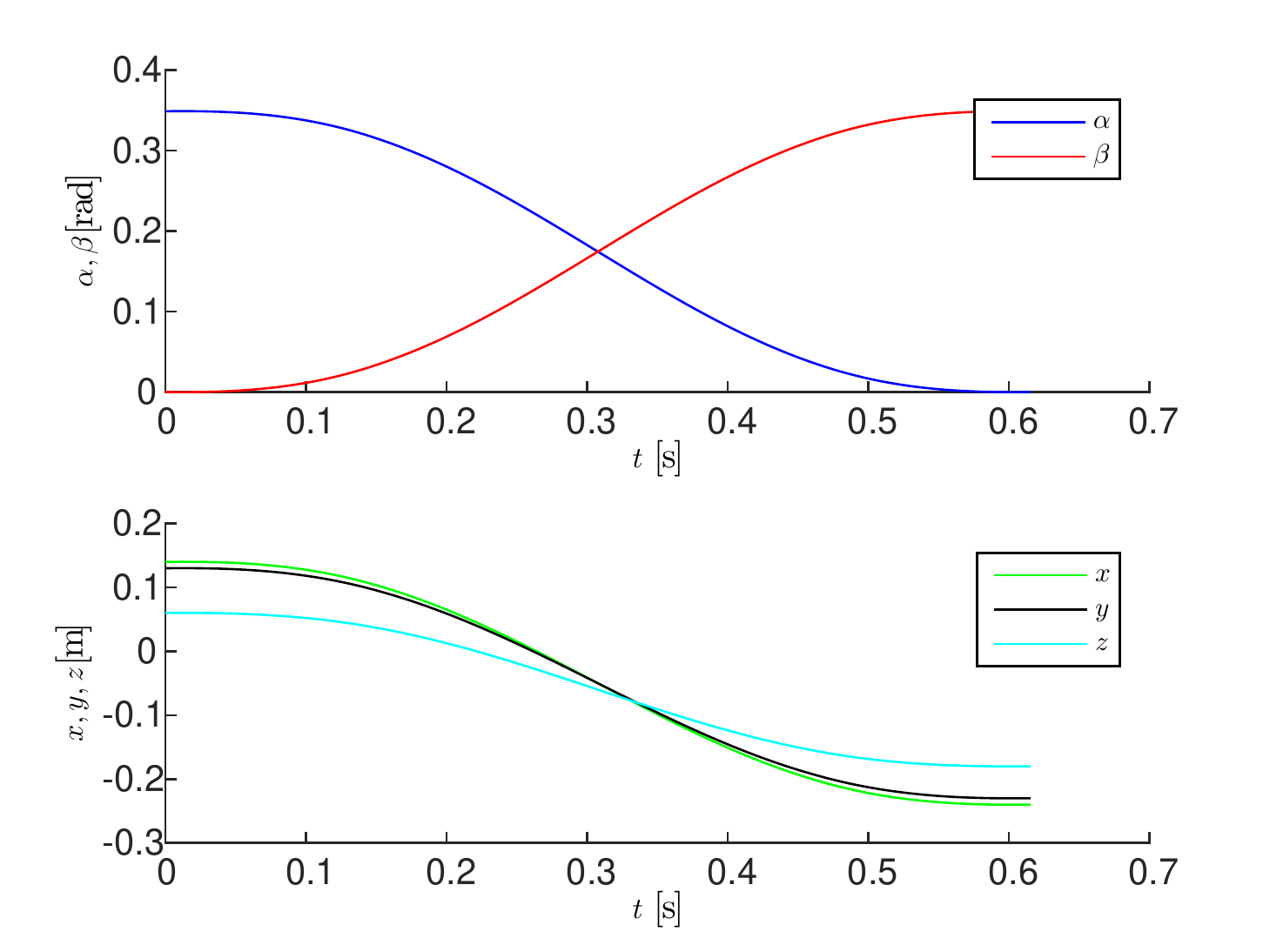} \\
				(b) $\bf{P}_2$ to $\bf{P}_3$
   \end{minipage} 
  \caption{Wrist angle and position of the end-effector along the linear trajectory}
  \protect\label{figure:Ligne_Cartesian_Position}
  \end{center}
\end{figure}

\begin{figure}[!ht]
  \begin{center}
	\begin{minipage}[c]{.49\linewidth}
        \includegraphics[scale=0.35]{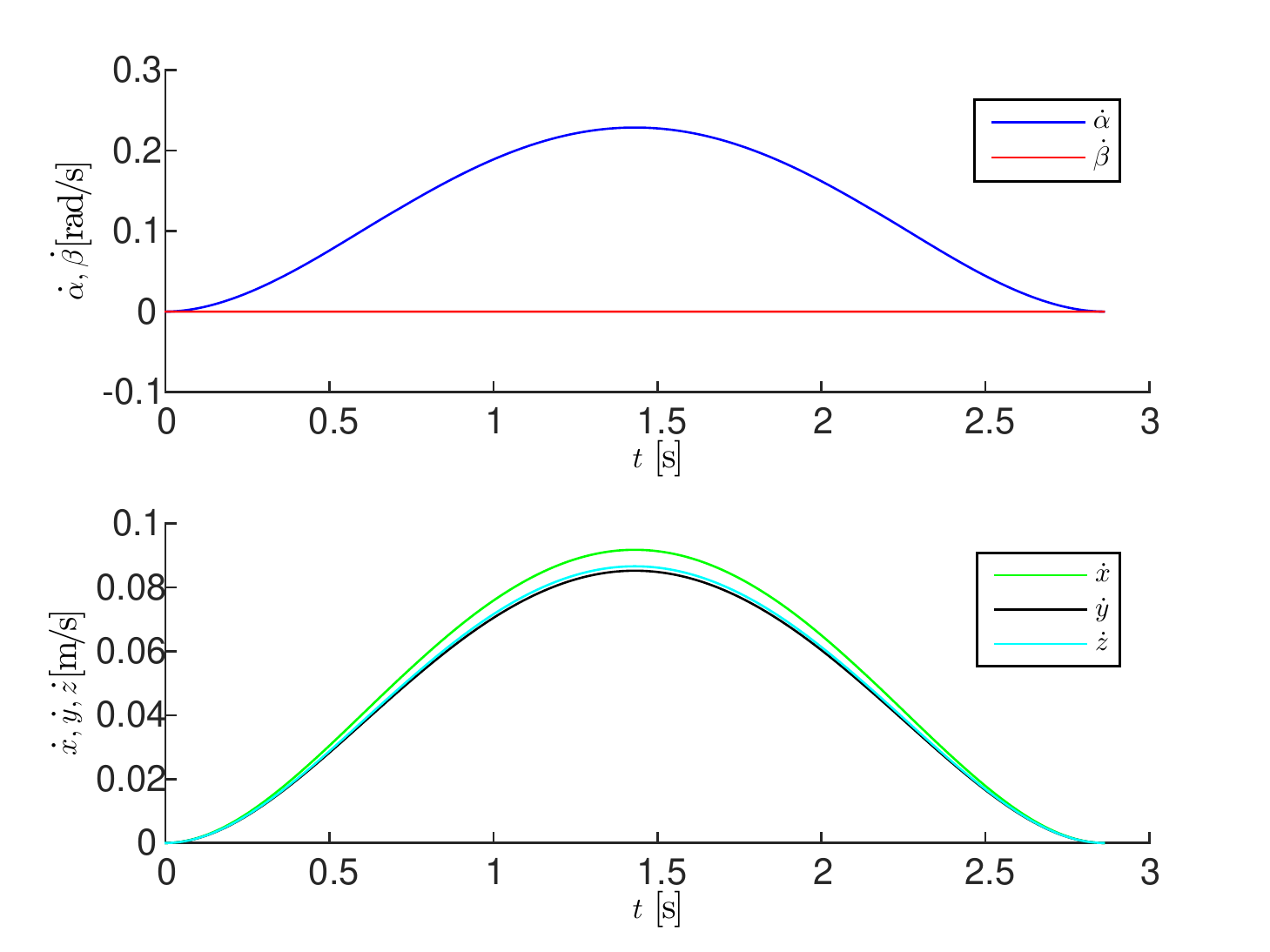}\\
				\small{(a) $\bf{P}_1$ to $bf{P}_2$}
   \end{minipage} \hfill
   \begin{minipage}[c]{.49\linewidth}        
        \includegraphics[scale=0.35]{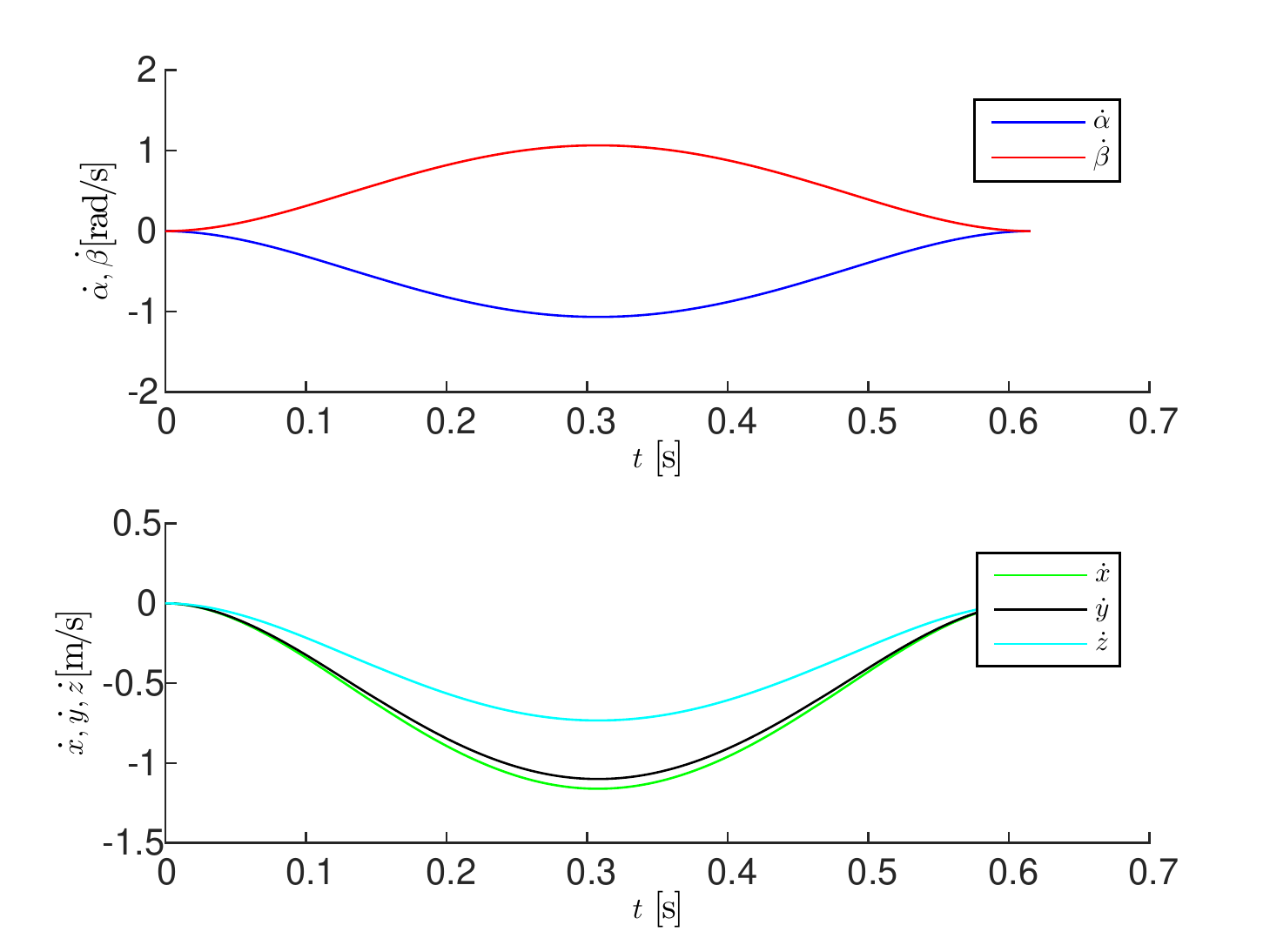}\\
				(b) $\bf{P}_2$ to $\bf{P}_3$
   \end{minipage} 
   \caption{Wrist angle velocity and velocity of the end-effector along the linear trajectory}
	 \protect\label{figure:Ligne_Cartesian_Vitesse}
 \end{center}
\end{figure}

\begin{figure}[!ht]
  \begin{center}
	\begin{minipage}[c]{.49\linewidth}
        \includegraphics[scale=0.35]{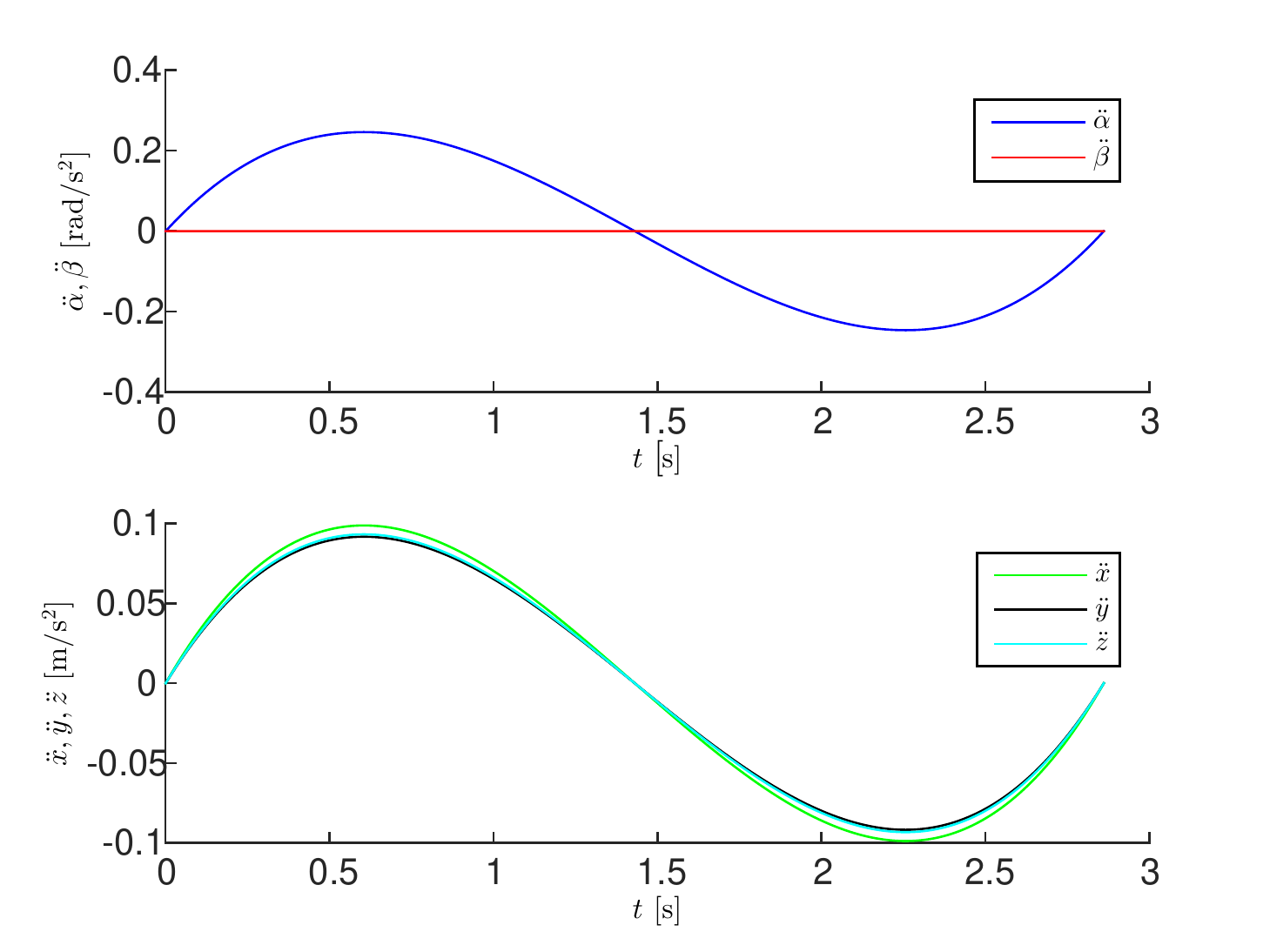} \\
				\small{(a) $\bf{P}_1$ to $\bf{P}_2$}
   \end{minipage} \hfill
   \begin{minipage}[c]{.49\linewidth}        
        \includegraphics[scale=0.35]{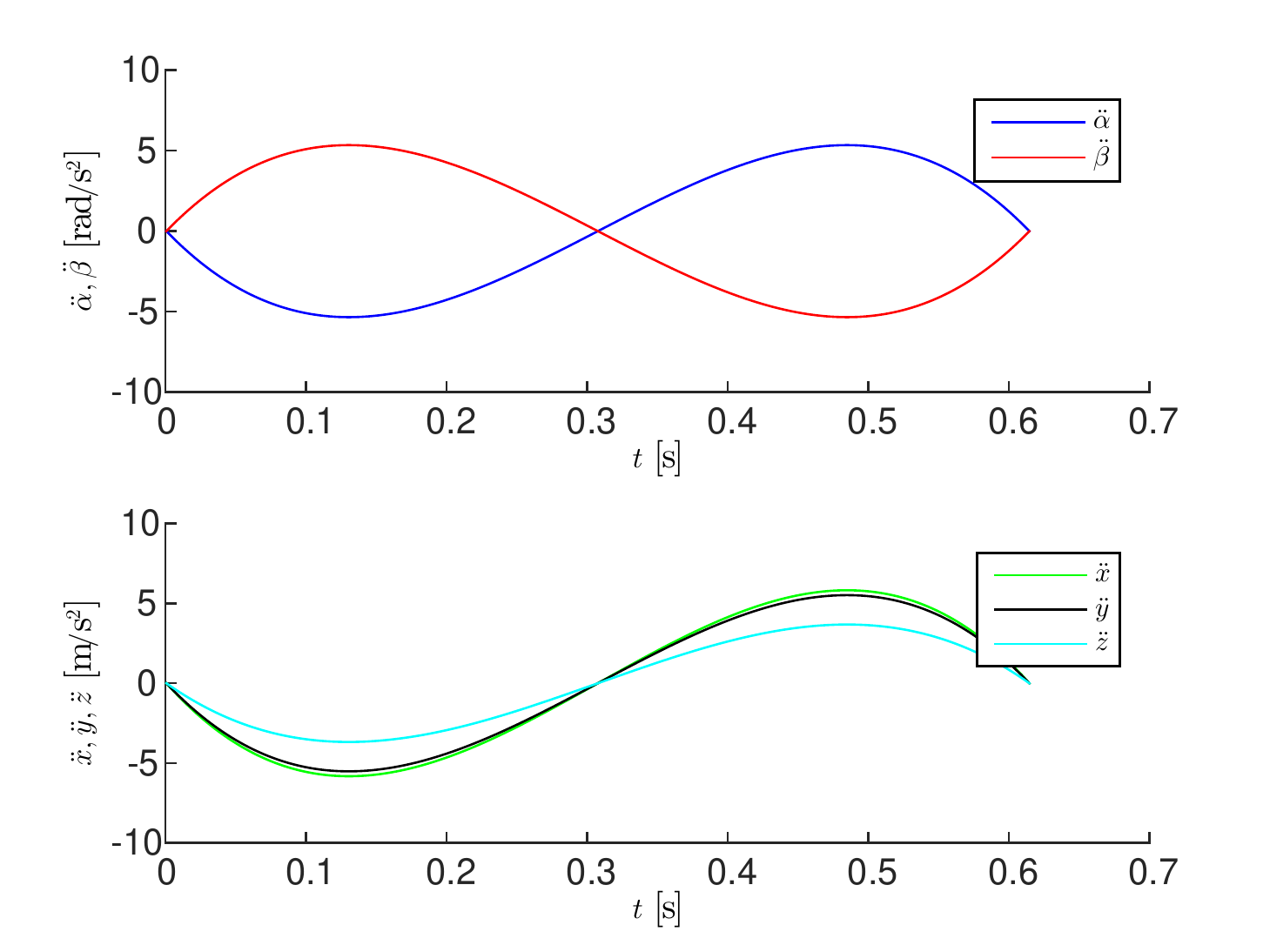} \\
				(b) $\bf{P}_2$ to $\bf{P}_3$
   \end{minipage} 
   \caption{Wrist angle acceleration and acceleration of the end-effector along the linear trajectory}
   \protect\label{figure:Ligne_Cartesian_Acceleration}
  \end{center}
\end{figure}

\begin{table}
\begin{center}
  \caption{Set of Orthoglide 5-axis end-effector poses used to define three linear trajectories}
	\label{table:ligne}
\begin{tabular}{ |c | c | c | c | c | }
 \hline			
   ~ & $\bf{P}_1$ & $\bf{P}_2$ & $\bf{P}_3$ & $\bf{P}_4$\\
	 \hline
   $\alpha$[degres] &   0  &  20 &   0 &  0\\
   $\beta$[degres]  &   0  &   0 &  20 &  0\\
	 $x$[mm]          &   0  & 140 &-240 &  0\\
	 $y$[mm]          &   0  & 130 &-230 &  0\\
	 $z$[mm]          & -72  &  60 &-180 &-72\\
 \hline  
 \end{tabular}
\end{center}
\end{table}
Figures~\ref{figure:Ligne_Cartesian_Position}, \ref{figure:Ligne_Cartesian_Vitesse} and \ref{figure:Ligne_Cartesian_Acceleration} show the position, velocity and acceleration profiles between $\bf{P}_1$ to $\bf{P}_2$ and $\bf{P}_2$ to $\bf{P}_3$ in the Cartesian space, respectively. Note that the trajectory between poses~$\bf{P}_3$ and~$\bf{P}_4$ is not depicted as it amounts to the one between poses~$\bf{P}_1$ and~$\bf{P}_2$.


We can observe that the maximum velocity is reached between poses $\bf{P}_3$ and $\bf{P}_4$ . It means that the projection in the joint space of the maximum speed is not greater than the maximum velocity or maximum acceleration able to be produced by the ball-screws. 

\begin{figure}[!ht]
  \begin{center}
        \includegraphics[scale=0.6]{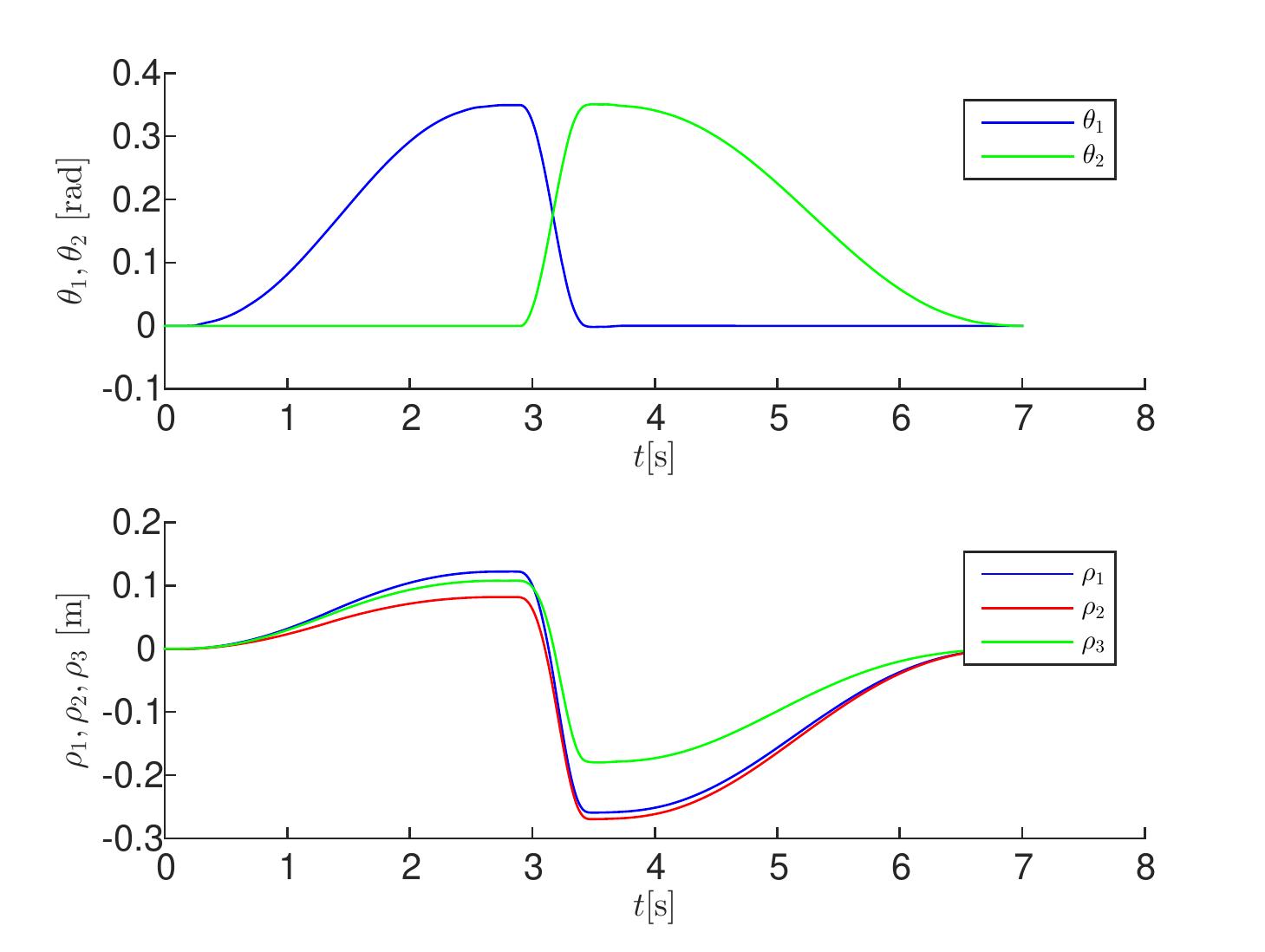}
        \caption{Joint values along the complete linear trajectory}
        \protect\label{figure:Ligne_Joint_Position}
  \end{center}
\end{figure}

\begin{figure}[!ht]
  \begin{center}
        \includegraphics[scale=0.6]{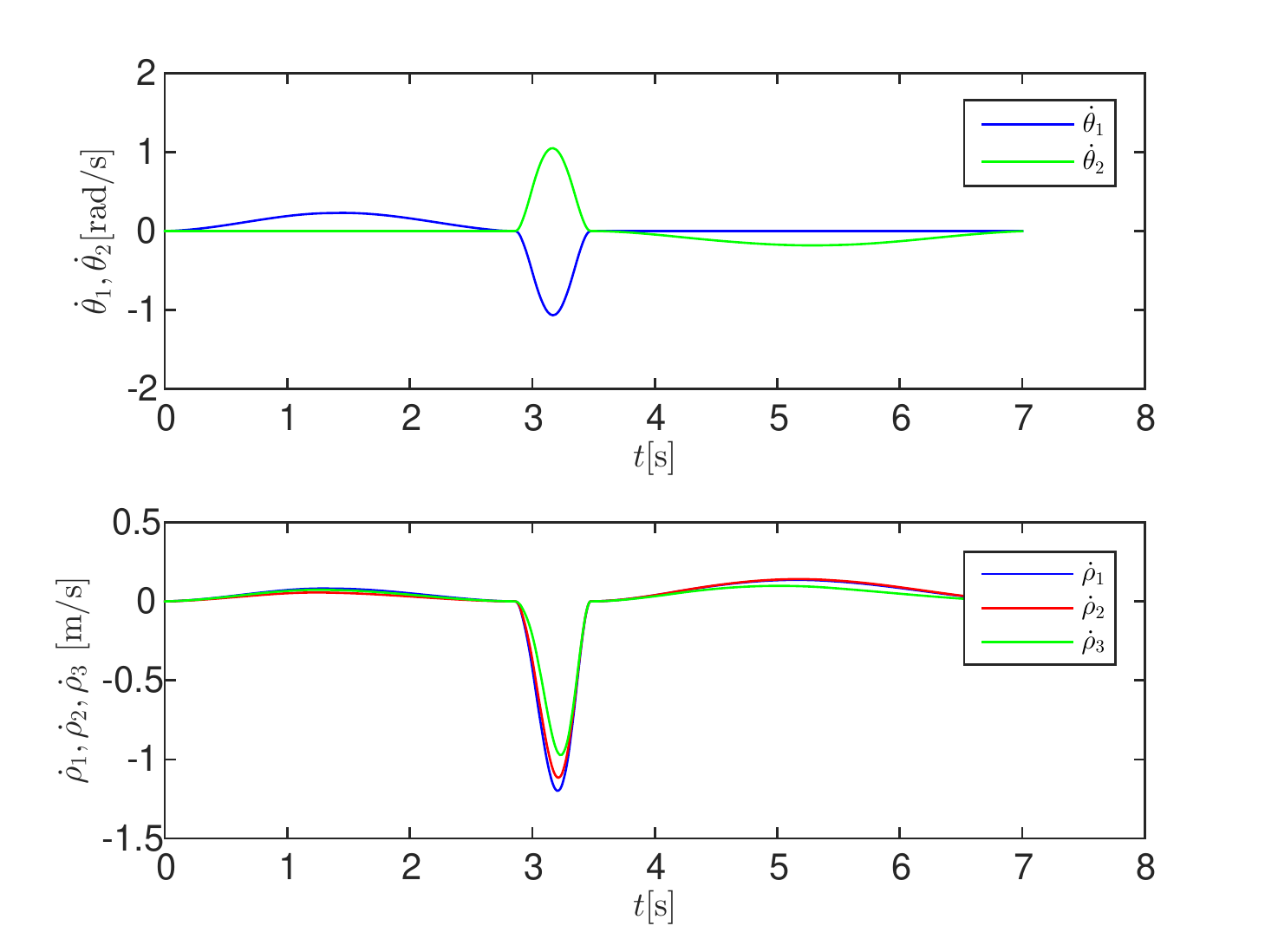}
        \caption{Joint velocities along the complete linear trajectory}
        \protect\label{figure:Ligne_Joint_Vitesse}
  \end{center}
\end{figure}

Figures~\ref{figure:Ligne_Joint_Position} and \ref{figure:Ligne_Joint_Vitesse}  show the joint positions and joint velocities along the three linear trajectories. 

\subsection*{Circular trajectory}
Table~\ref{table:cercle} describes the path associated with a circular trajectory, the geometric center of the circle being point~$P_C$ and its radius is equal to~$R$. Poses~$\bf{P}_2$ and ${\bf P}_3$ are added before and after the circular trajectory, respectively. These poses are defined by the radius of the circle, the two angles~$\eta_{min}$ and~$\eta_{max}$ define both the ends of the arc of circle. Thus, from pose ${\bf P}_1$ to pose $\bf{P}_2$ and from pose $\bf{P}_3$ to pose $\bf{P}_4$, a linear trajectory is generated. The orientation of the circular path is characterized by two angles $\alpha_1$ and $\beta_1$ about the $z$ and $y$ axes, respectively. $A_2$ and $B_2$, ($A_3$ and $B_3$, resp.) define the orientation of the tool tip for poses~$\bf{P}_2$ and~$\bf{P}_3$. For the illustrative example,  $R=30$mm and $A_2=20$, $B_2=0$, $A_3=0$ and $B_3=20$ degrees.  
\begin{table}
\begin{center}
  \caption{Set of Orthoglide 5-axis end-effector poses used to define two linear trajectories and one circular trajectory}
	\label{table:cercle}
\begin{tabular}{ |c | c | c | c |  }
 \hline			
   ~ & $\bf{P}_1$ & $\bf{P}_C$ & $\bf{P}_4$ \\
	 \hline
   $\alpha$[degres] &   0  &  0 &   0\\
   $\beta$[degres]  &   0  &  0 &   0\\
	 $x$[mm]          &   0  & 10 &   0\\
	 $y$[mm]          &   0  & 10 &   0\\
	 $z$[mm]          & -72  & 10 & -72\\
 \hline  
 \end{tabular}
\end{center}
\end{table}

The local coordinate vector~$\bf{X}_l$ of the end-effector is defined as:
\begin{equation}
\bf{X}_l= \left[
             \begin{array}{c}
						x_l \\
				    y_l \\
						z_l
						 \end{array}
          \right]
					=
					\left[
             \begin{array}{c}
						R \cos(\eta_{min}+\Delta \eta r(t)) \\
				    R \sin(\eta_{min}+\Delta \eta r(t)) \\
						0
						 \end{array}
          \right]
\end{equation}
Its global coordinate vector $\bf{X}_g$ is defined as:
\begin{equation}
  \bf{X}_g= \bf{P}_C + \bf{R}(\alpha_1) \bf{R}(\beta_1) \bf{X}_l
\end{equation}
The pose of the tool is then defined as:
\begin{equation}
  \bf{P}(t)= \left[\begin{array}{c}
	                   A_{2} + (A_3-A_2) r(t)\\
										 B_{2} + (B_3-B_2) r(t) \\
										 x_C + x_g \\
										 y_C + y_g \\
										 z_C + z_g
	                 \end{array}
	           \right]
\end{equation}
The local velocity is: 
\begin{equation}
\dot{\bf{X}}_l= \left[
             \begin{array}{c}
						\dot{x}_l \\
				    \dot{y}_l \\
						\dot{z}_l
						 \end{array}
          \right]
					=
					\left[
             \begin{array}{c}
						-R \sin(\eta_{min}+\Delta \eta r(t)) \dot{r}(t) \Delta \eta\\
				     R \cos(\eta_{min}+\Delta \eta r(t)) \dot{r}(t) \Delta \eta\\
						0
						 \end{array}
          \right]
\end{equation}
And its velocity $\dot{\bf{X}}_g$ in the global reference frame is
\begin{equation}
  \dot{\bf{X}}_g=  \bf{R}(\alpha_1) \bf{R}(\beta_1) \dot{\bf{V}}_l
\end{equation}
The velocity is then defined as
\begin{equation}
  \bf{V}(t)= \left[\begin{array}{c}
	                   (A_3-A_2) \dot{r}(t)\\
										 (B_3-B_2) \dot{r}(t) \\
										 \dot{x}_g \\
										 \dot{y}_g \\
										 \dot{z}_g
	                 \end{array}
	           \right]
\end{equation}
Finally, the local acceleration is 
\begin{eqnarray}
\ddot{\bf{X}}_l= \left[
             \begin{array}{c}
						\ddot{x}_l \\
				    \ddot{y}_l \\
						\ddot{z}_l
						 \end{array}
          \right] 
					=
					\left[
             \begin{array}{c}
						-R( (\dot{r}(t) \Delta \eta)^2 \cos(\gamma) + \ddot{r}(t) \Delta \eta \sin(\gamma)\\
				     R(-(\dot{r}(t) \Delta \eta)^2 \sin(\gamma) + \ddot{r}(t) \Delta \eta \cos(\gamma) \\
						0
						 \end{array}
          \right]
\end{eqnarray}
with $\gamma= \eta_{min}+\Delta \eta r(t)$. And the acceleration $\ddot{\bf{X}}_g$ in the global reference frame is
\begin{equation}
  \ddot{\bf{X}}_g=  \bf{R}(\alpha_1) \bf{R}(\beta_1) \dot{\bf{A}}_l
\end{equation}
The acceleration is then defined as
\begin{equation}
  \bf{A}(t)= \left[\begin{array}{c}
	                   (A_3-A_2) \ddot{r}(t)\\
										 (B_3-B_2) \ddot{r}(t) \\
										 \ddot{x}_g \\
										 \ddot{y}_g \\
										 \ddot{z}_g
	                 \end{array}
	           \right]
\end{equation}

\begin{figure}
  \begin{center}
	\begin{minipage}[c]{.49\linewidth}
        \includegraphics[scale=0.35]{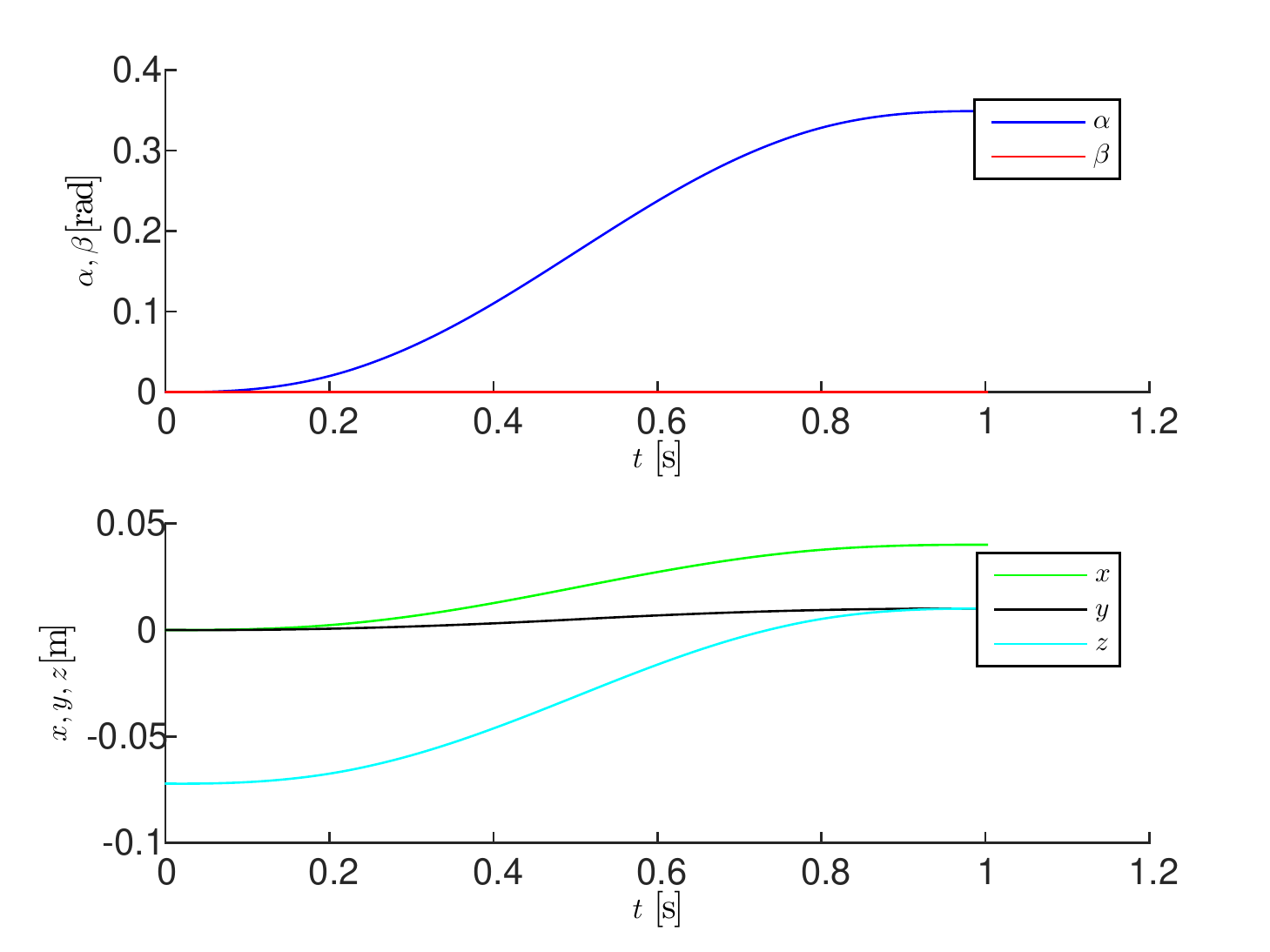} \\
				\small{(a) $\bf{P}_1$ to $\bf{P}_2$}
   \end{minipage} \hfill
   \begin{minipage}[c]{.49\linewidth}        
        \includegraphics[scale=0.35]{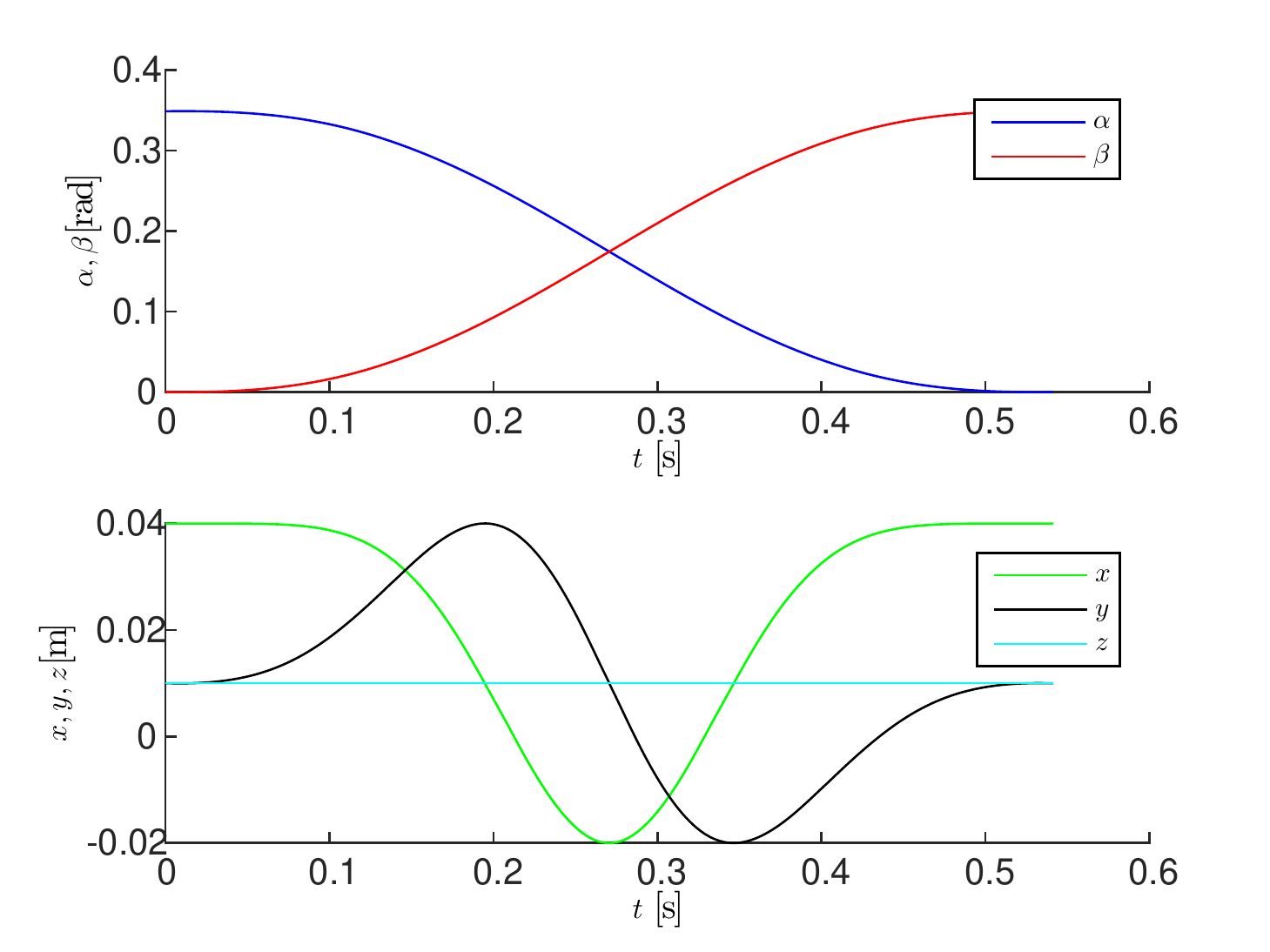} \\
				(b) $\bf{P}_2$ to $\bf{P}_3$
   \end{minipage} 
  \protect\label{figure:Cercle_2_Cartesian_Position}
  \caption{Wrist angle and position of the end-effector along the circular trajectory}
  \end{center}
\end{figure}

\begin{figure}
  \begin{center}
	\begin{minipage}[c]{.49\linewidth}
        \includegraphics[scale=0.35]{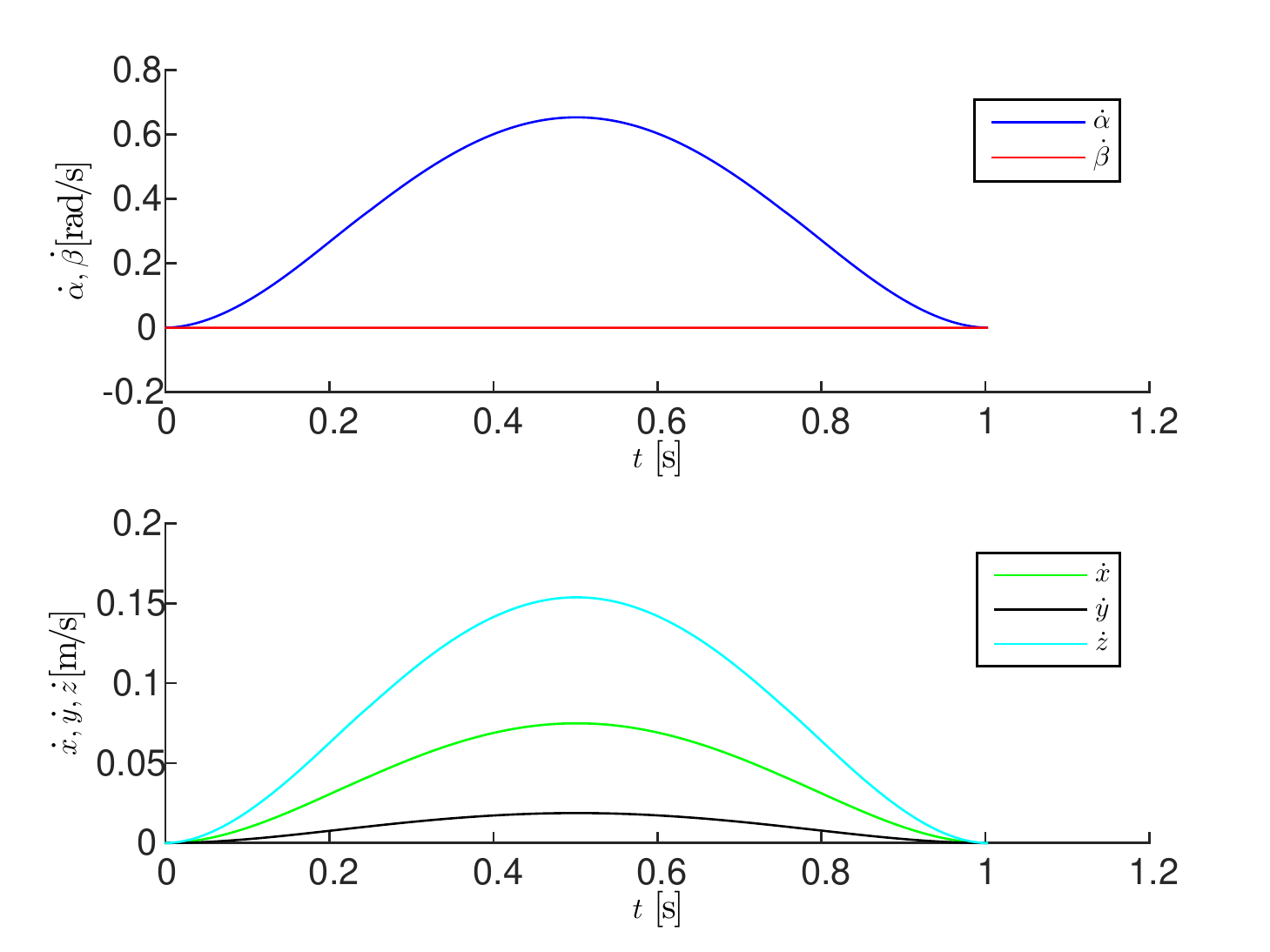}\\
				\small{(a) $\bf{P}_1$ to $\bf{P}_2$}
   \end{minipage} \hfill
   \begin{minipage}[c]{.49\linewidth}        
        \includegraphics[scale=0.35]{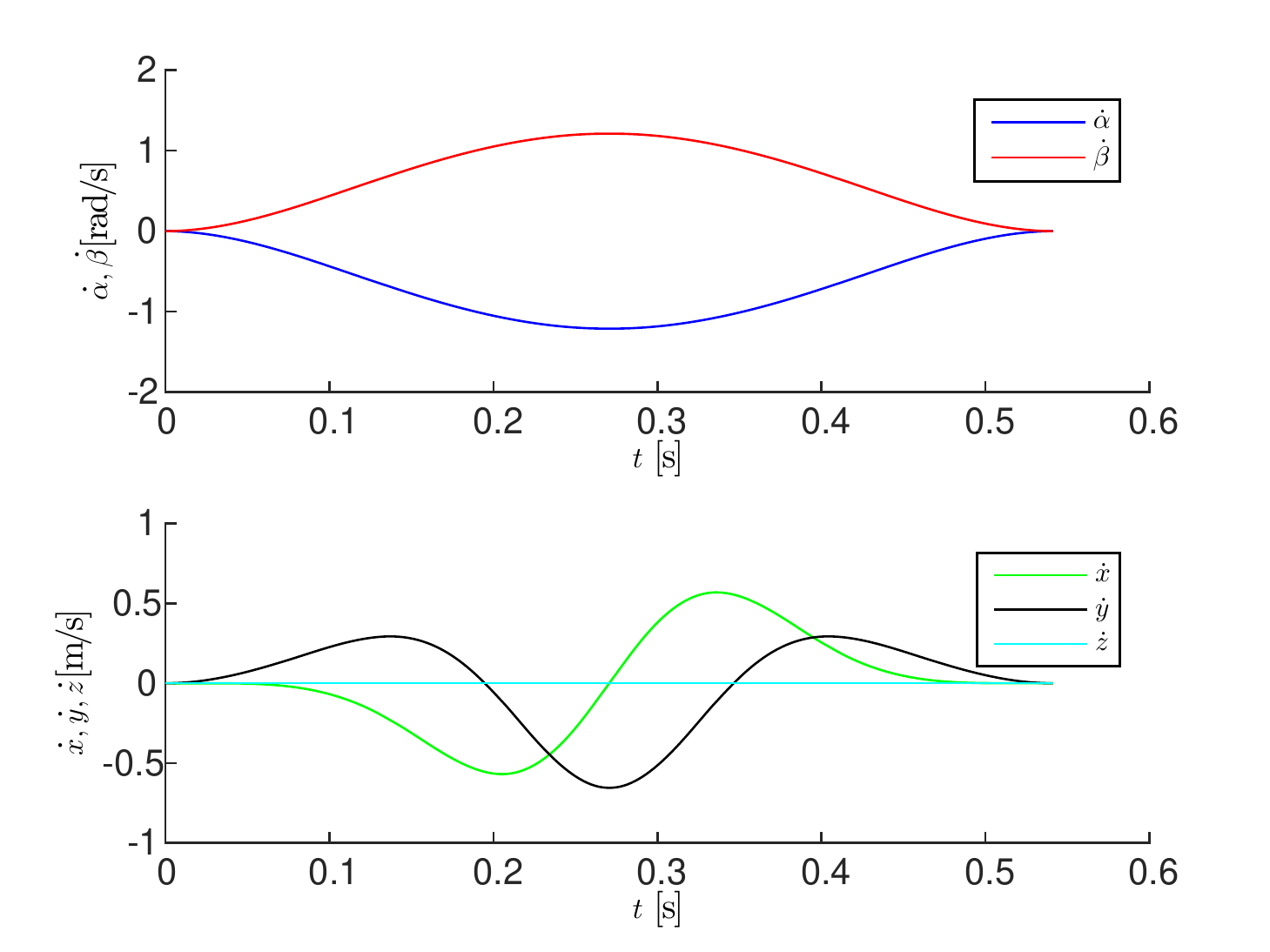}\\
				(b) $\bf{P}_2$ to $\bf{P}_3$
   \end{minipage} 
	 \protect\label{figure:Cercle_2_Cartesian_Vitesse}
   \caption{Wrist angle velocity and velocity of the end-effector along the circular trajectory}
 \end{center}
\end{figure}
\begin{figure}
  \begin{center}
	\begin{minipage}[c]{.49\linewidth}
        \includegraphics[scale=0.35]{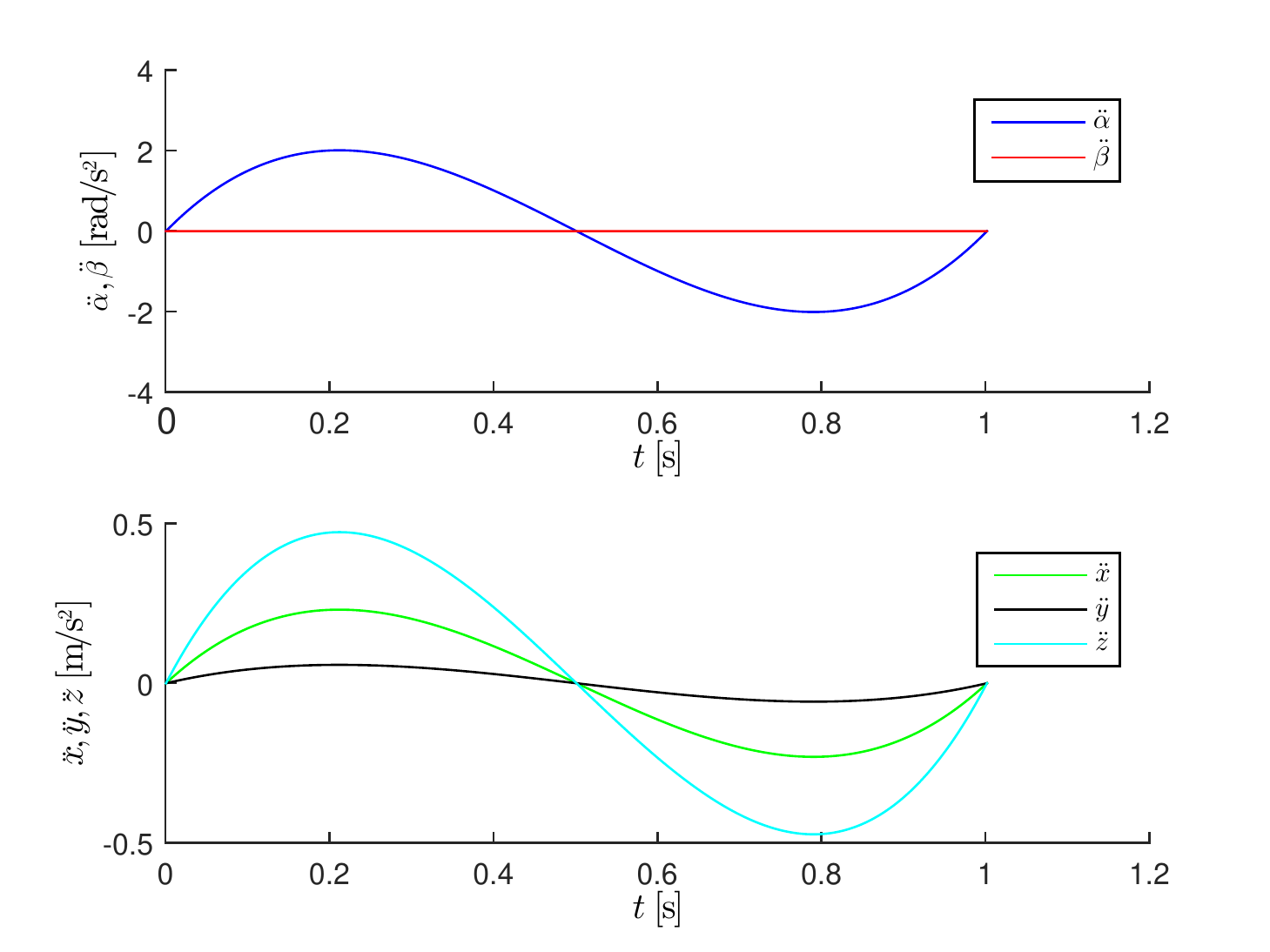} \\
				\small{(a) $\bf{P}_1$ to $\bf{P}_2$}
   \end{minipage} \hfill
   \begin{minipage}[c]{.49\linewidth}        
        \includegraphics[scale=0.35]{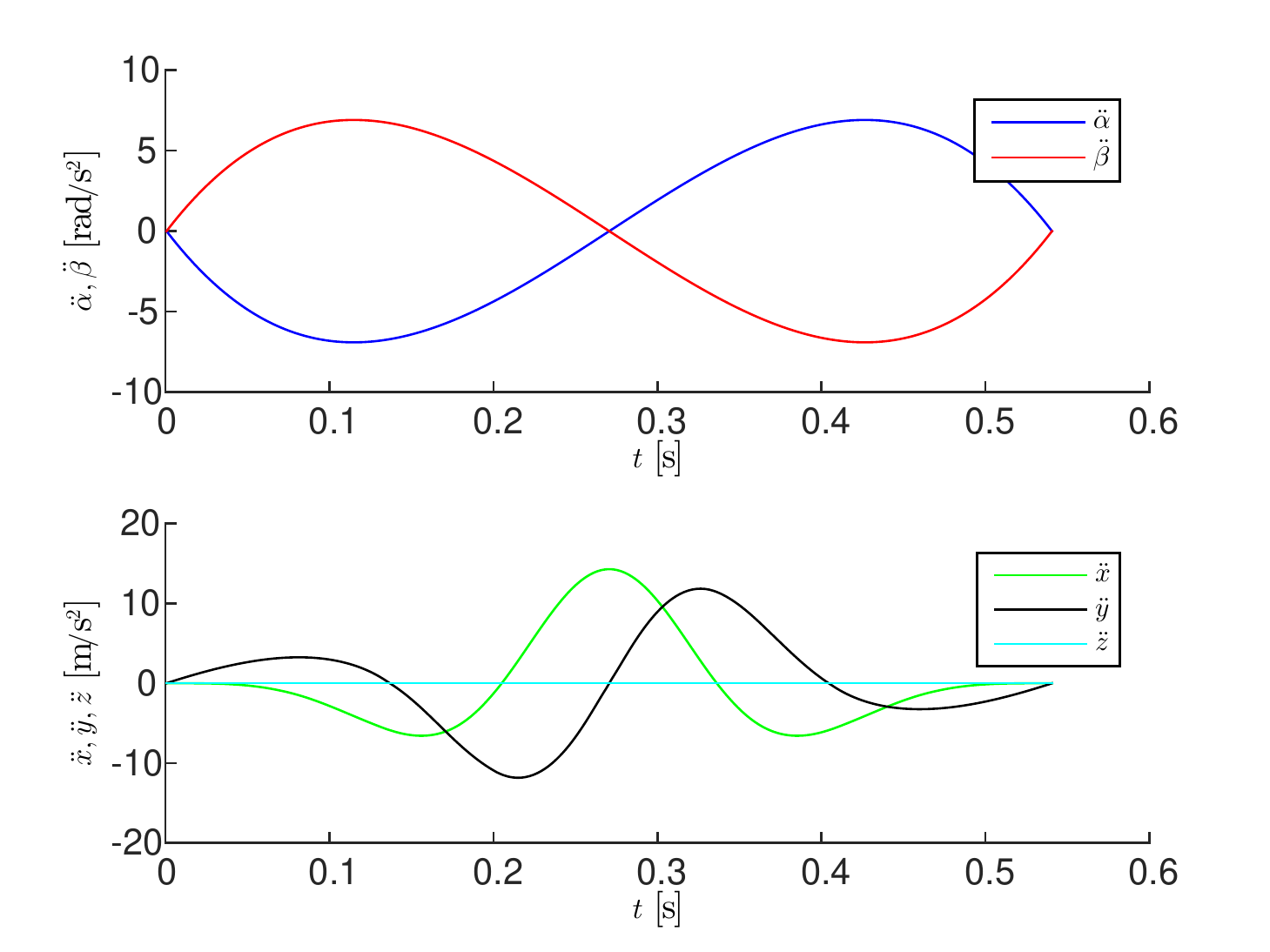} \\
				(b) $\bf{P}_2$ to $\bf{P}_3$
   \end{minipage} 
   \protect\label{figure:Cercle_2_Cartesian_Acceleration}
   \caption{Wrist angle acceleration and acceleration of the end-effector along the circular trajectory}
  \end{center}
\end{figure}

\begin{figure}
  \begin{center}
        \includegraphics[scale=0.6]{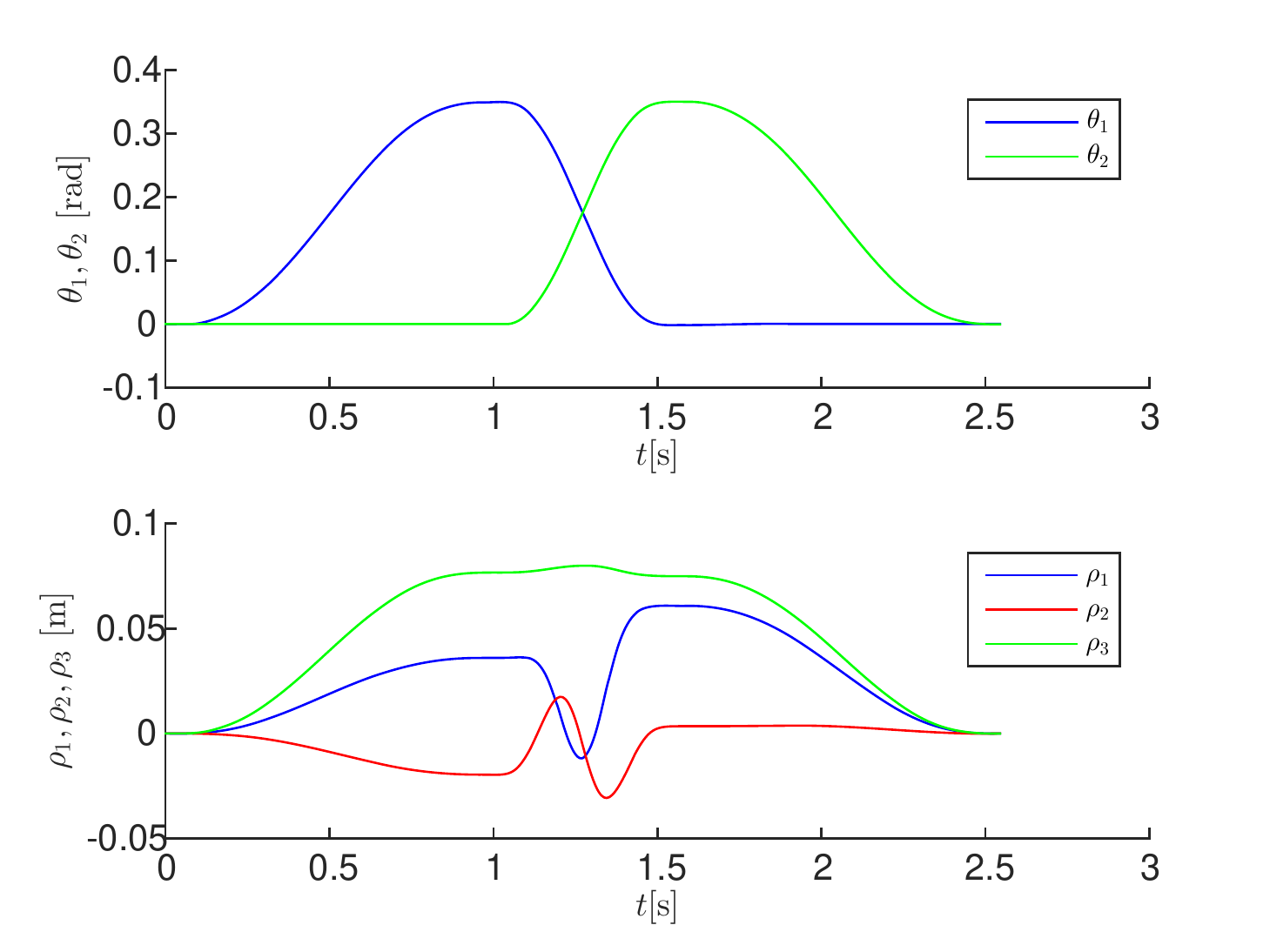}
        \caption{Joint values along the complete circular trajectory}
        \protect\label{figure:Cercle_Joint_Position}
  \end{center}
\end{figure}

\begin{figure}
  \begin{center}
        \includegraphics[scale=0.6]{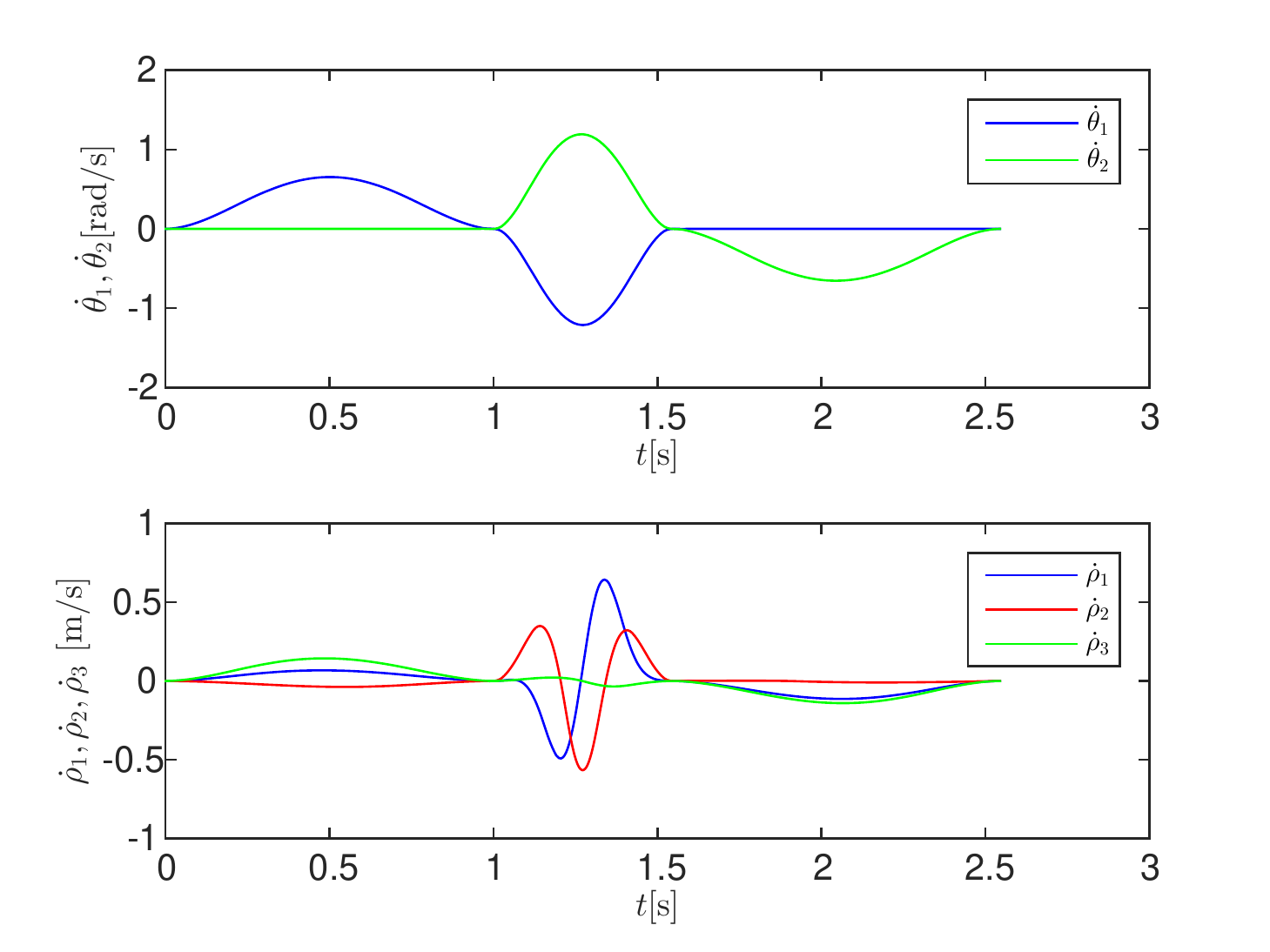}
        \caption{Joint velocities along the complete circular trajectory}
        \protect\label{figure:Cercle_Joint_Vitesse}
  \end{center}
\end{figure}

Figures~\ref{figure:Cercle_Joint_Position} and \ref{figure:Cercle_Joint_Vitesse} show the joint positions and joint velocities for the complete circular trajectory. During the circular trajectory, the maximum joint acceleration is reached and led to a maximum speed reduction. In that case, the execution time of the complete trajectory is increased. Then, the coefficient: $\gamma_V$ which is the ratio of the maximum joint velocity to the speed of the associated motor and $\gamma_A$ which is the ratio of the maximum joint acceleration to the acceleration of the associated motor are computed. As a result, the traveling time becomes larger and the maximum Cartesian velocity decreases up to 0.6~m/s. 
\begin{equation}
t_f= t_f \max(\gamma_V,~\sqrt{\gamma_A})
\end{equation}
\subsection*{Evaluation of the Orthoglde 5-axis performance}
\begin{figure}
  \begin{center}
        \includegraphics[scale=0.6]{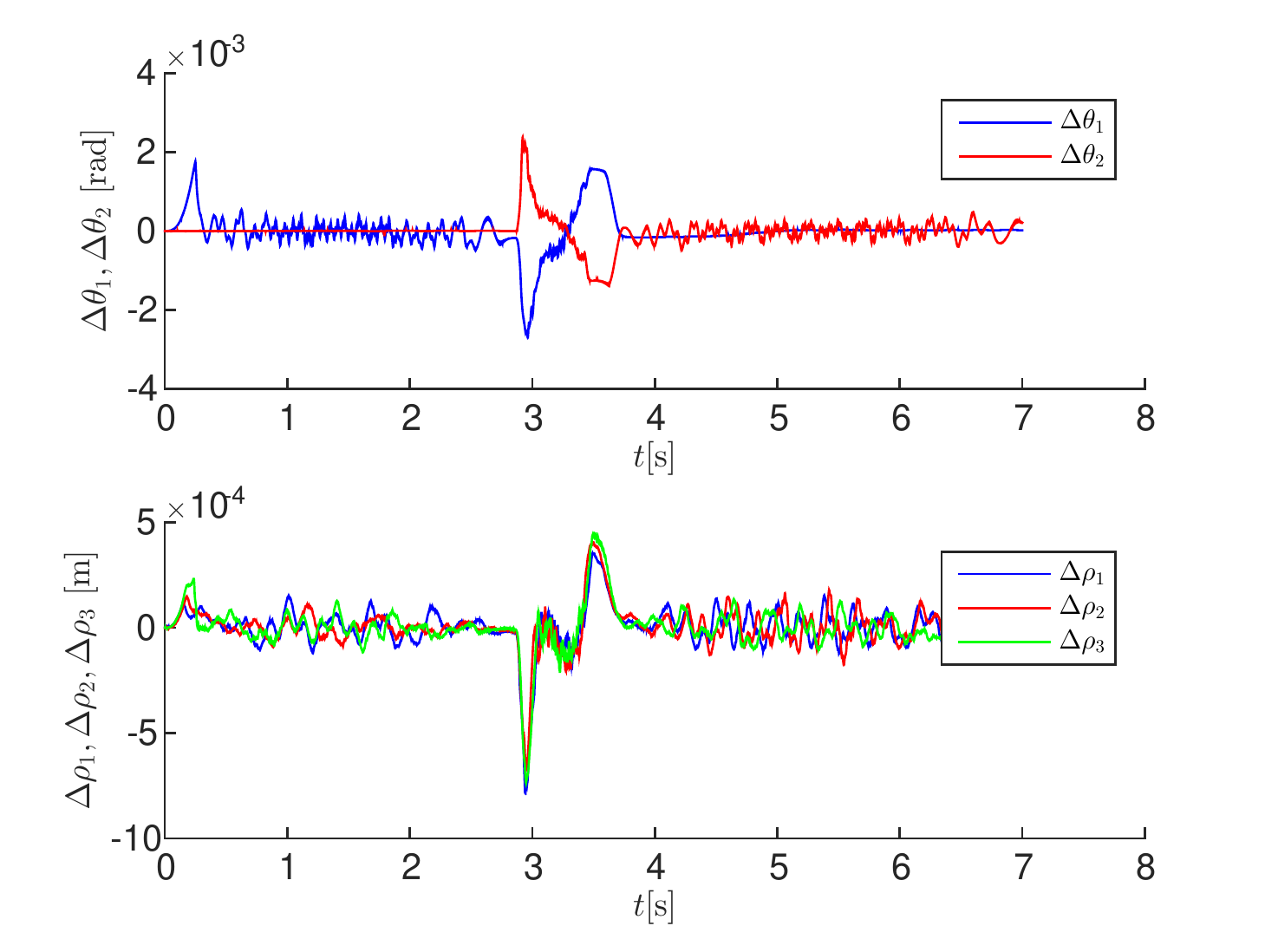}
        \caption{Joint angles and position errors along the complete linear trajectory}
        \protect\label{figure:Ligne_Joint_Erreur}
  \end{center}
\end{figure}

\begin{figure}
  \begin{center}
        \includegraphics[scale=0.6]{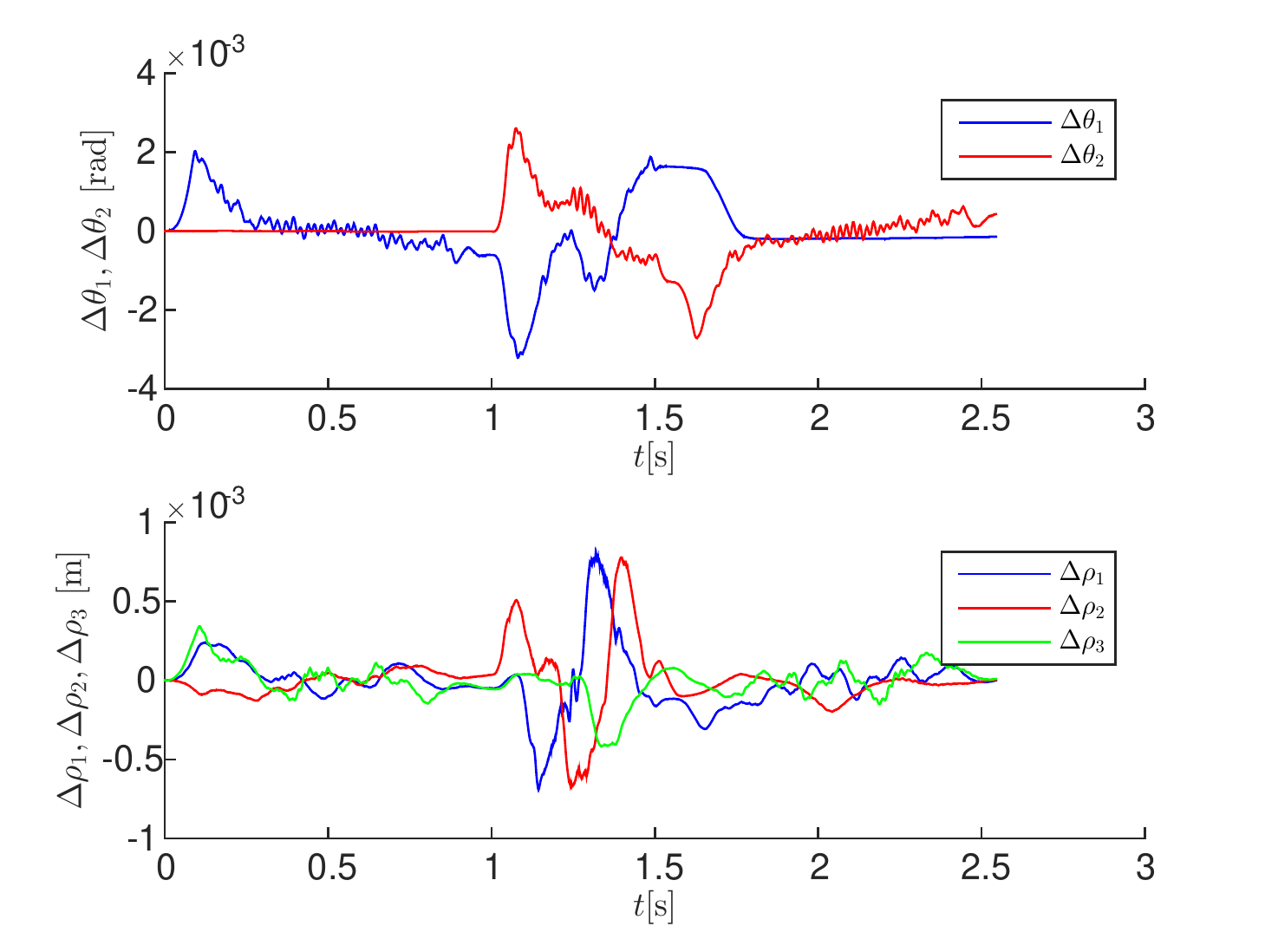}
        \caption{Joint angles and position errors along the complete circular trajectory}
        \protect\label{figure:Cercle_Joint_Erreur}
  \end{center}
\end{figure}

Figures~\ref{figure:Ligne_Joint_Erreur} and \ref{figure:Cercle_Joint_Erreur} show the joint position errors along the complete trajectories. It appears that a simple control loop is enough when the velocity is low as the corresponding maximum error in the actuated prismatic lengths is equal to 0.2~mm. The maximum error is reached during the linear trajectory $0.7$ mm, when the maximum velocity is reached. 

\begin{figure}
  \begin{center}
	\begin{minipage}[c]{.49\linewidth}
        \includegraphics[scale=0.35]{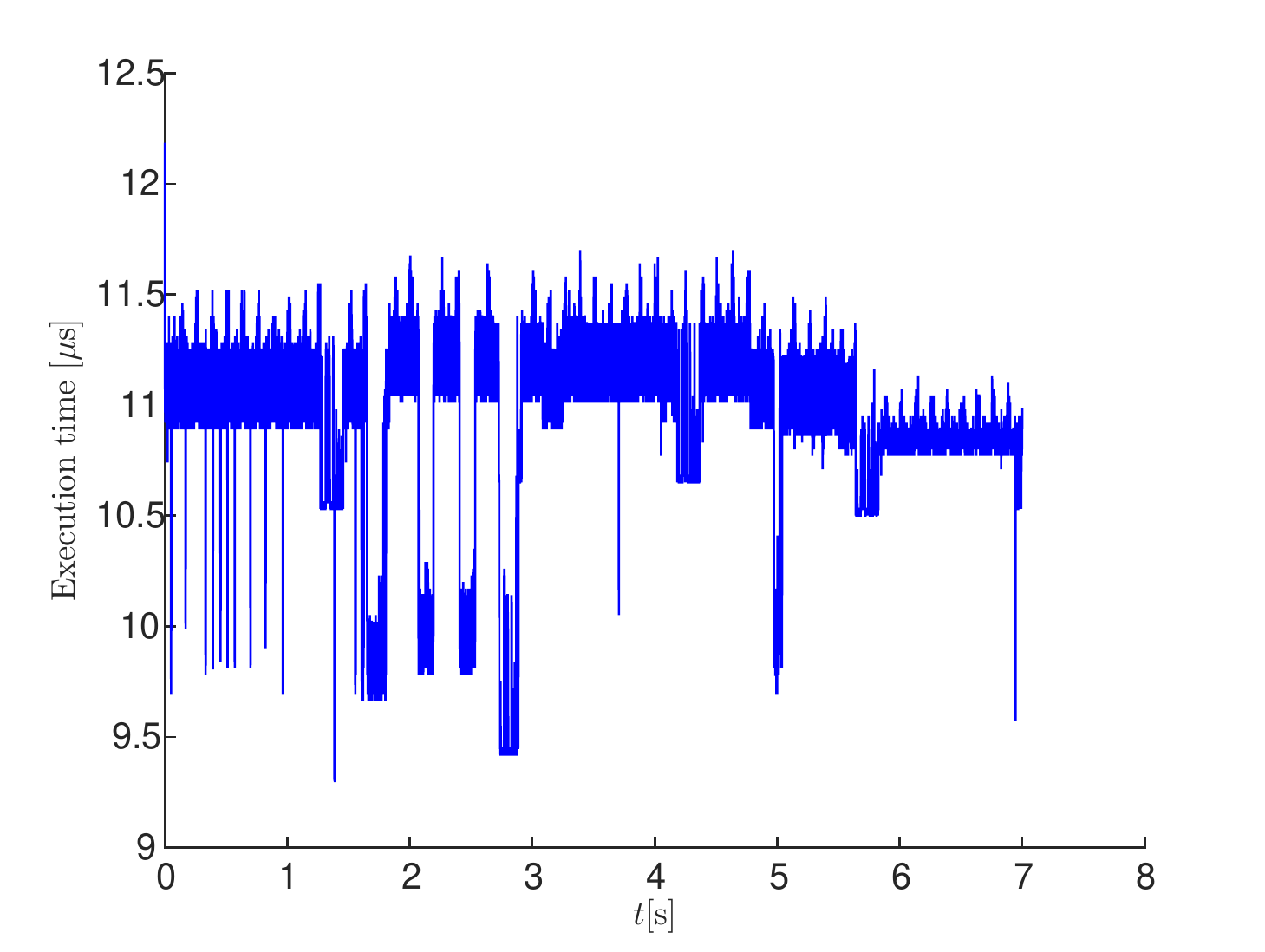} \\
				\small{(a)}
   \end{minipage} \hfill
   \begin{minipage}[c]{.49\linewidth}        
        \includegraphics[scale=0.35]{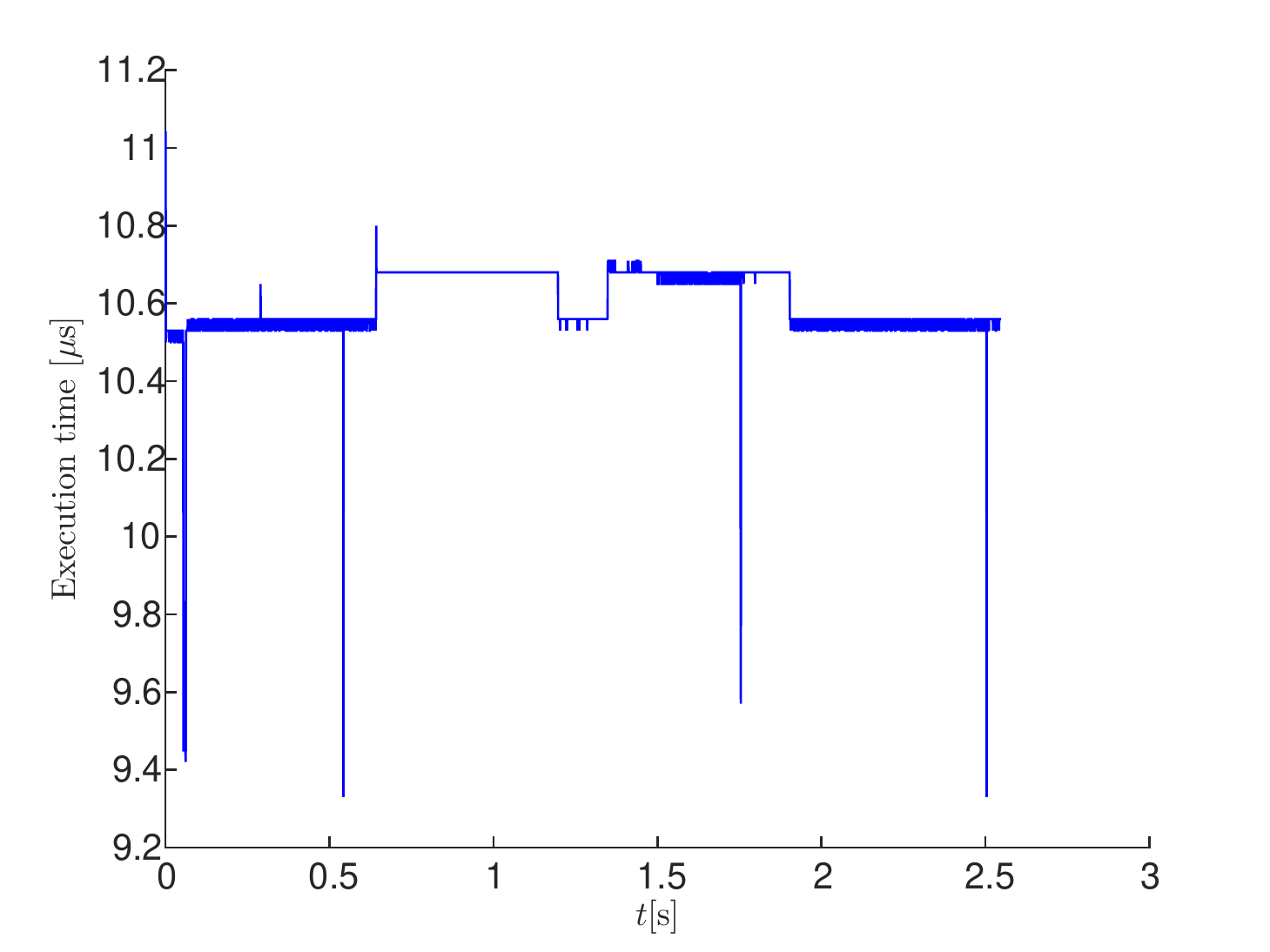} \\
				\small{(b)}
   \end{minipage} 
   \caption{Execution time for (a) the linear trajectory and (b) the circular trajectory}
   \protect\label{figure:Temps}
  \end{center}
\end{figure}

\begin{figure}
  \begin{center}
	\begin{minipage}[c]{.49\linewidth}
        \includegraphics[scale=0.35]{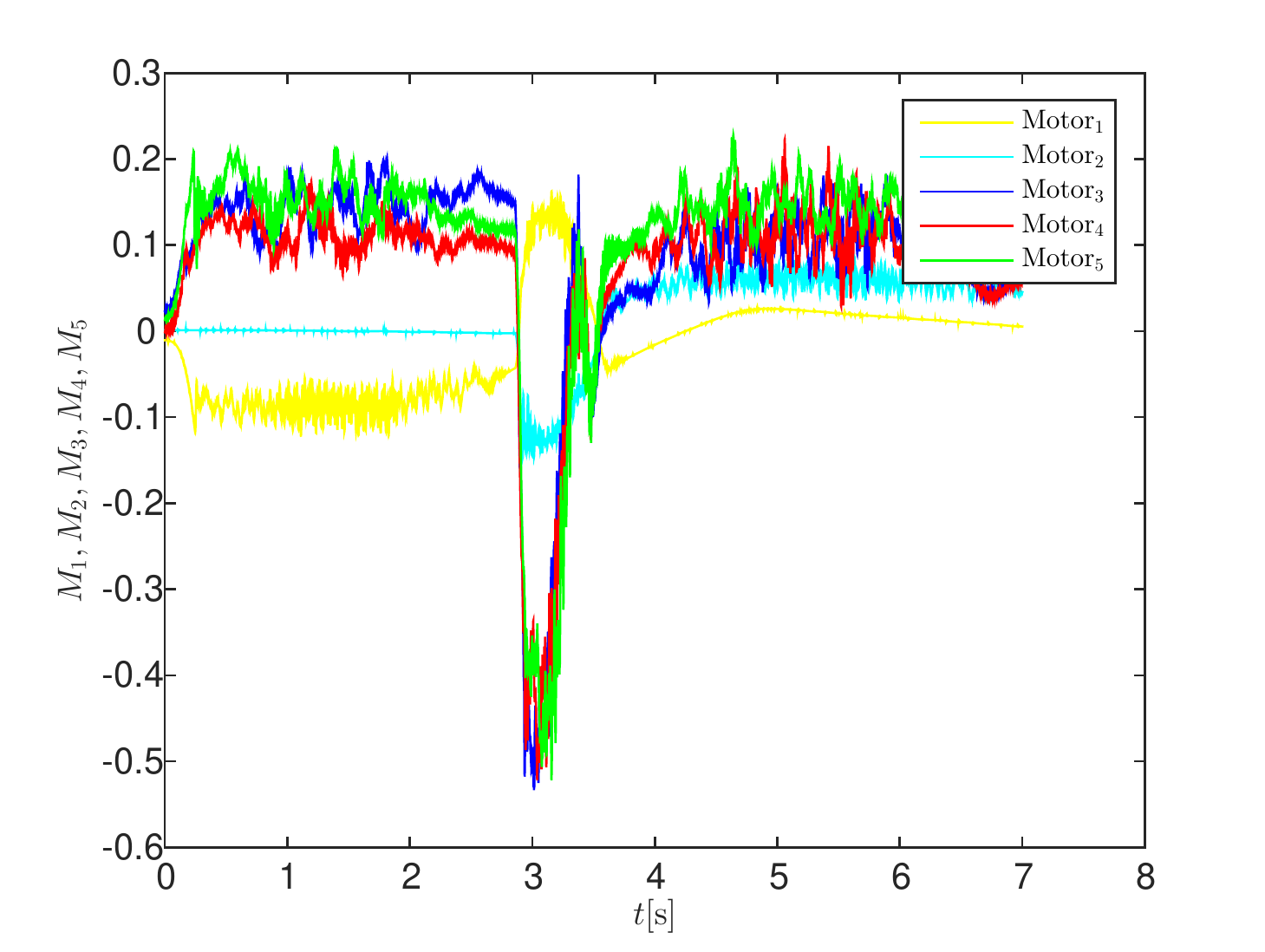} \\
				\small{(a)}
   \end{minipage} \hfill
   \begin{minipage}[c]{.49\linewidth}        
        \includegraphics[scale=0.35]{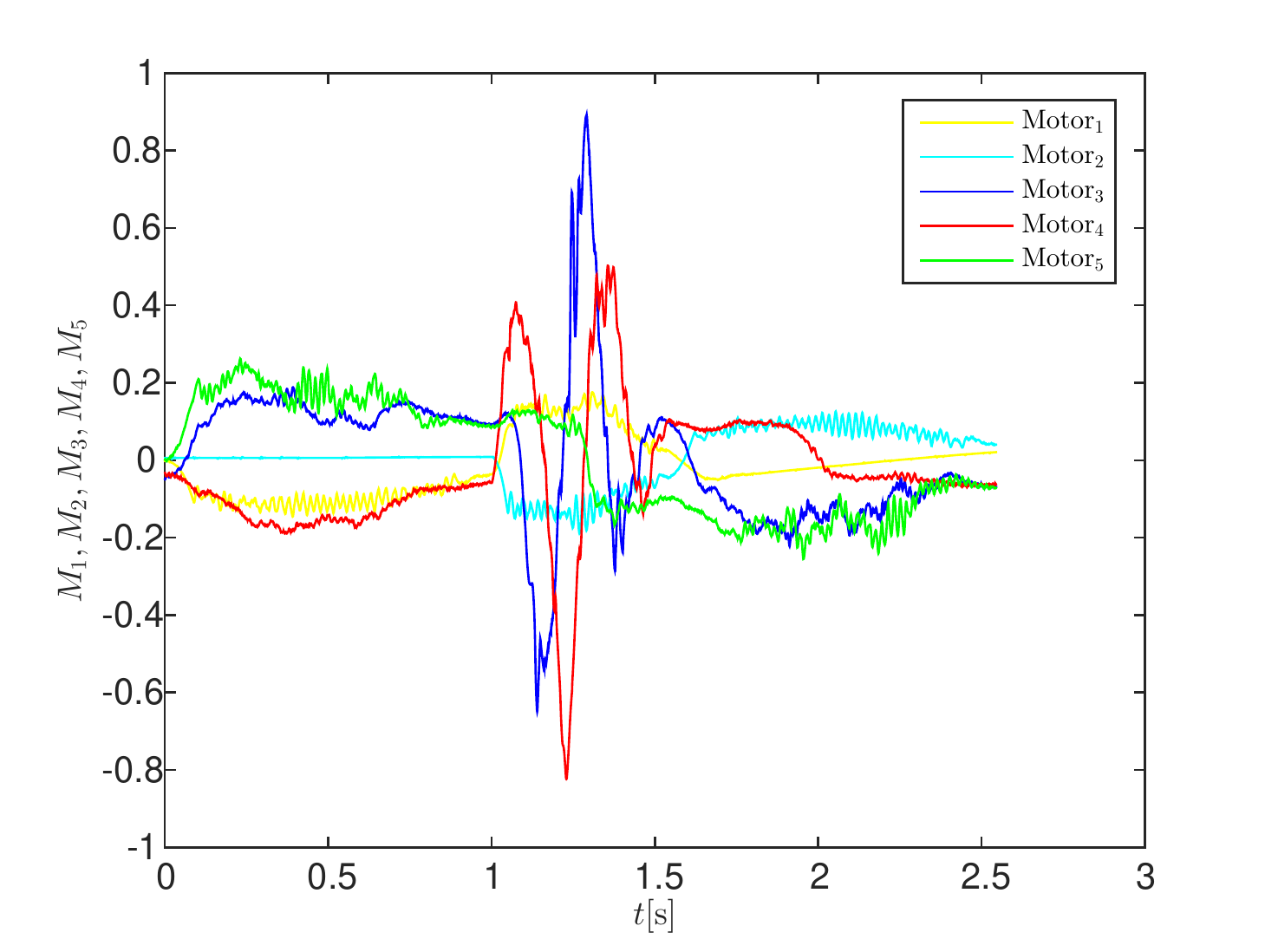} \\
				\small{(b)}
   \end{minipage} 
   \caption{Percentage of the motor torques sent to the drives (a) the linear trajectory and (b) the circular trajectory}
   \protect\label{figure:Commande}
  \end{center}
\end{figure}

Figure~\ref{figure:Temps} gives the execution time for both trajectories. We can notice that the time is stable and close to 10 $\mu$s. This time includes the computation of the velocity, the direct kinematics and othe control loop. As we have a real-time operating system, we have also data to send to the computer in which we have the Control Desk software able to make data acquisitions.
Figure~\ref{figure:Commande} shows the control signal sent to the drive. The variation is between -1 and 1, which amount to the maximum continuous torques that each motor may produce. For an acceleration equal to 1.3~G, it is 80\% of the maximum motor torque for the~$3^{rd}$ and~$4^{th}$ motors. As a consequence, we expect to increase the efficiency of the control loop later on by using a part of peak torques in order to reach higher accelerations. 
\section*{CONCLUSIONS}
This paper dealt with the kinematic modeling of the Orthoglide 5-axis and its use for a computed torque control of the semi-industrial prototype. From the semi-industrial prototype developed at IRCCyN, we studied the use of this model for generating movements. We have shown that unlike most of the parallel kinematics machines, the kinematics of the Orthoglide 5-axis is rather simple.
It was shown that the expressions of the geometric models and inverse Jacobian matrices are simple to express. The assembly of two parallel robots leads to an inverse kinematic Jacobian matrix composed of three blocks, the inverse kinematic Jacobian matrix of each robot and one additional block corresponding to the coupling between the two parallel robots. 
For performing computed torque control, we have generated a set combining positions, velocities and accelerations of the actuators that are sent to the control loop. Two types of trajectories were studied and validated experimentally.  We have shown that for linear motions of up to 1 m / s, the tracking error in the actuator position is lower than one millimeter. 
Experiments are ongoing to refine the estimate of the masses and inertia in motion and to identify the friction parameters.
\begin{acknowledgment}
The work presented in this paper was partially funded by the Region ``Pays de la Loire'', the European projects NEXT (``Next Generation of Productions System'') and by the FEDER Robotic Project of IRCCyN.
\end{acknowledgment}
 

\begin{thebibliography}{12}
\bibitem{Clavel:1988}
R. Clavel, `` DELTA, a Fast Robot with Parallel Geometry,'' Pro. Of the 18th Int. Symp. of Robotic Manipulators, IFR Publication, pp. 91-100, 1988.
\bibitem{Merlet:2006}
J-P. Merlet, ``Parallel robots,'' Springer, 2nd edition, ISBN 978-1-4020-4133-4, 2006.
\bibitem{Wenger:2000}
P. Wenger, D. Chablat, ``Kinematic Analysis of a New Parallel Machine Tool: the Orthoglide,'' 7th International Symposium on Advances in Robot Kinematics, Slovenia, June 2000.
\bibitem{Ur-Rehman:2008} 
R. Ur-Rehman, S. Caro, D. Chablat, P. Wenger, ``Kinematic and Dynamic Analyses of the Orthoglide 5-axis,'' 7th France-Japan/ 5th Europe-Asia Congress on Mechatronics, Le Grand-Bornand, France, May 21-23, 2008.
\bibitem{Gosselin:1994}
C. Gosselin, and J.F. Hamel, ``The agile eye: a high performance three-degree-of-freedom camera-orienting device,'' IEEE Int. conference on Robotics and Automation, San Diego, 1994, pp. 781-787.
\bibitem{Gosselin:2004}
C. Gosselin, X. Kong, ``Cartesian parallel manipulators,''  US Patent  6729202 B2, May 4, 1994.
\bibitem{Caron:1997}
F. Caron, ``Analyse et conception d'un manipulateur parall\`ele sph\'erique à deux degr\`es de libert\'e pour l'orientation d'une cam\'ra,'' M.Sc., Universit\'e Laval, Qu\'ebec, 1997.
\bibitem{Chablat:2003}
D. Chablat, and P. Wenger, ``Architecture Optimization, of a 3-DOF Parallel Mechanism for Machining Applications, The Orthoglide,'' IEEE Trans. on Robotics and Automation 19, 2003, pp.403–410.
\bibitem{Chablat:2006}
D. Chablat and P. Wenger, ``Device for the movement and orientation of an object in space and use thereof in rapid machining,'' European Patent  1597017, Sept. 6, 2006.
\bibitem{Khalil:2002}
W. Khalil, E. Dombre, ``Modeling, Identification and Control of Robots,''  HPS, Butterworth Heinemann, ISBN 9781903996133, 2002.
\bibitem{Dspace:2015}
Dspace, ``dSPACE - DS1103 PPC Controller Board'', from www.dspace.com/en/pub/home/products/hw/singbord/ \goodbreak ppcconbo.cfm, Retrieved January 21, 2015.
\bibitem{Pashkevich:2006}
A. Pashkevich, D. Chablat and P. Wenger, ``Kinematics and Workspace Analysis of a Three-Axis Parallel Manipulator: the Orthoglide'', Robotica, Volume 24, Issue 1, pp. 39-49, January 2006.
\bibitem{Pashkevich:2005}
A. Pashkevich, P. Wenger and D.Chablat, ``Design Strategies for the Geometric Synthesis of Orthoglide-type Mechanisms,'' Journal of Mechanism and Machine Theory, Vol. 40(8), 2005, pp. 907-930.
\bibitem{Pashkevich:2008}
A. Pashkevich, D. Chablat and P. Wenger, ``Stiffness Analysis of 3-d.o.f. Overconstrained Translational Parallel Manipulators,'' Proc. IEEE Int. Conf. Rob. and Automation, May 2008.
\bibitem{Chablat:2006b}
D. Chablat, and P. Wenger, ``A six degree of freedom Haptic device based on the Orthoglide and a hybrid Agile Eye,'' 30th Mechanisms \& Robotics Conference (MR),  Philadelphia, USA, 2006.
 \end{thebibliography}
\end{document}